\def\paperID{}
\def\confName{CVPR}
\def\confYear{2025}

\def\paperTitle{PatchAlign3D: Local Feature Alignment for Dense 3D Shape Understanding}

\def\authorBlock{
Souhail Hadgi$^{1}$ \qquad
Bingchen Gong$^{1}$ \qquad
Ramana Sundararaman$^{1}$ \qquad
Emery Pierson$^{1}$ \\
Lei Li$^{2}$ \qquad
Peter Wonka$^{3}$ \qquad
Maks Ovsjanikov$^{1}$ \\[0.4em]
$^{1}$\'Ecole polytechnique \qquad
$^{2}$University of Virginia \qquad
$^{3}$KAUST 
}

\newif\ifreview 
\newif\ifarxiv \newcommand{\arxiv}{\arxivtrue}
\newif\ifcamera 
\newif\ifrebuttal 

\arxiv 

\pdfoutput=1
\documentclass[10pt,twocolumn,letterpaper]{article}
\ifreview \usepackage[review]{cvpr} \fi
\ifarxiv \usepackage[pagenumbers]{cvpr} \fi
\ifrebuttal \usepackage[rebuttal]{cvpr} \fi
\ifcamera \usepackage{cvpr} \fi

\usepackage{graphicx}	
\usepackage{amsmath}	
\usepackage{amssymb}	
\usepackage{booktabs}
\usepackage{times}
\usepackage{microtype}
\usepackage{epsfig}
\usepackage{caption}
\usepackage{float}
\usepackage{placeins}
\usepackage{color, colortbl}
\usepackage{stfloats}
\usepackage{enumitem}
\usepackage{tabularx}
\usepackage{xstring}
\usepackage{multirow}
\usepackage{xspace}
\usepackage{url}
\usepackage{subcaption}
\usepackage{xcolor}
\usepackage[hang,flushmargin]{footmisc}
\usepackage[table]{xcolor} 
\usepackage{graphicx}      
\usepackage{capt-of}  
\definecolor{OursTint}{RGB}{232,246,255} 
\definecolor{PosGreen}{RGB}{0,120,0}
\definecolor{NegRed}{RGB}{170,0,0}
\ifcamera \usepackage[accsupp]{axessibility} \fi

\usepackage{adjustbox}
\usepackage{array} 
% ...

\definecolor{darkgreen}{RGB}{0,100,0}

\ifarxiv  \fi

\newcommand{\R}[1]{{%
    \textbf{%
        \ifstrequal{#1}{1}{\textcolor{red}{R#1}}{%
        \ifstrequal{#1}{2}{\textcolor{blue}{R#1}}{%
        \ifstrequal{#1}{3}{\textcolor{magenta}{R#1}}{%
        \ifstrequal{#1}{4}{\textcolor{teal}{R#1}}{%
                           \textcolor{cyan}{R#1}%
        }}}}%
    }%
}}

\usepackage{xr-hyper}

\makeatletter
\newcommand*{\addFileDependency}[1]{
  \typeout{(#1)}
  \@addtofilelist{#1}
  \IfFileExists{#1}{}{\typeout{No file #1.}}
}

\makeatother
\newcommand*{\myexternaldocument}[1]{
    \externaldocument{#1}
    \addFileDependency{#1.tex}
    \addFileDependency{#1.aux}
}

\definecolor{cvprblue}{rgb}{0.21,0.49,0.74}
\usepackage[pagebackref,breaklinks,colorlinks,allcolors=cvprblue]{hyperref}
\usepackage[capitalize]{cleveref}
\crefname{section}{Sec.}{Secs.}
\crefname{table}{Table}{Tables}
\crefname{figure}{Fig.}{Figs.}

\ifarxiv \crefname{appendix}{App.}{Apps.}
\else \crefname{appendix}{Suppl.}{Suppls.} \fi

\unless\ifarxiv \myexternaldocument{_supplementary} \fi

\begin{document}
\title{\paperTitle}
\author{\authorBlock}

\twocolumn[{
  \renewcommand\twocolumn[1][]{#1}%
  \maketitle

  \begin{center}
    \captionsetup{type=figure} 
    \includegraphics[width=\linewidth]{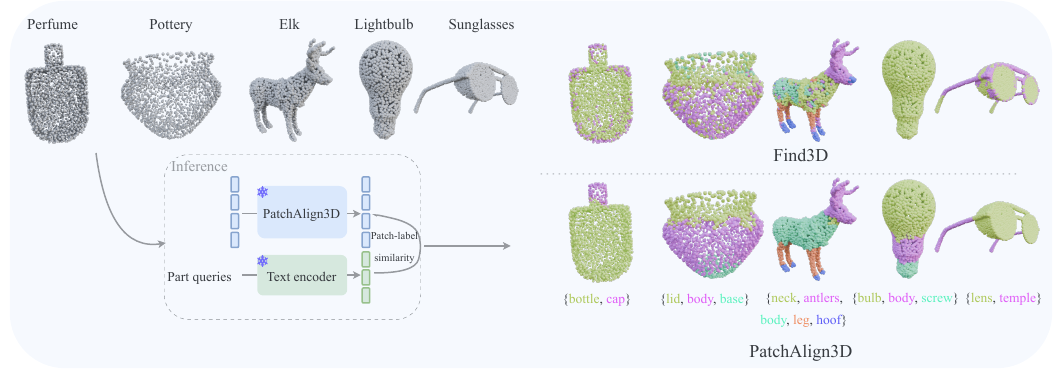}
    \caption{
    PatchAlign3D is a point cloud transformer encoder that produces language-aligned 
    patch-level features. 
    Through two-stage training,
    it enables zero-shot
    3D part segmentation from simple text queries in a feed-forward manner. Compared to prior methods such as Find3D \cite{ma2025find}, PatchAlign3D yields sharper, more accurate, and less noisy segmentation boundaries, paving the way for local 3D foundation models.}
    \label{fig:teaser}
  \end{center}
}]  

\newcommand{\mypara}[1]{\vspace{1ex}\noindent\textbf{#1}~}

\begin{abstract}
Current foundation models for 3D shapes excel at \emph{global} tasks (retrieval, classification) but transfer poorly to \emph{local} part-level reasoning. Recent approaches leverage vision and language foundation models to directly solve dense tasks through multi-view renderings and text queries. While promising, these pipelines require expensive inference over multiple renderings, depend heavily on large language-model (LLM) prompt engineering for captions, and fail to exploit the inherent 3D geometry of shapes. We address this gap by introducing an encoder-only 3D model that produces language-aligned patch-level features directly from point clouds. Our pre-training approach builds on existing data engines that generate part-annotated 3D shapes by pairing multi-view SAM regions with VLM captioning. Using this data, we train a point cloud transformer encoder in two stages: (1) distillation of dense 2D features from visual encoders such as DINOv2 into 3D patches, and (2) alignment of these patch embeddings with part-level text embeddings through a multi-positive contrastive objective. Our 3D encoder achieves zero-shot 3D part segmentation with fast single-pass inference without any test-time multi-view rendering, while significantly outperforming previous rendering-based and feed-forward approaches across several 3D part segmentation benchmarks. \noindent\textbf{Project website:} \href{https://souhail-hadgi.github.io/patchalign3dsite}{souhail-hadgi.github.io/patchalign3dsite}

\end{abstract}
\section{Introduction}
\label{sec:intro}

Understanding 3D geometry is a fundamental challenge in computer vision, central to fields ranging from robotics and AR/VR to content creation and scientific analysis.

Recent efforts to build 3D foundation models~\cite{xue2023ulip, liu2023openshape, zhou2023uni3d} have achieved impressive results on \emph{global} tasks like retrieval and classification, surpassing by a large margin VLM-based approaches such as PointCLIPv2 \cite{zhu2022pointclip}, which rely solely on 2D renderings for 3D understanding. However, open-vocabulary dense prediction tasks, such as 3D part segmentation, remain largely dominated by multi-view pipelines~\cite{zhu2022pointclip, liu2023partslip, abdelreheem2023satr, garosi20253d}. These methods render each 3D shape into multiple images, extract features using powerful 2D vision models like DINOv2~\cite{oquab2023dinov2} and CLIP~\cite{radford2021learning}, and then fuse the resulting 2D predictions back into 3D. This paradigm effectively transfers informative 2D semantic priors to 3D, enabling open-vocabulary 3D part understanding, and has therefore become the \emph{de facto} approach for dense 3D tasks.

Despite being effective, multi-view pipelines lack geometric grounding, as their predictions are primarily based on 2D appearance cues rather than the underlying 3D structure. Their inference process is also computationally expensive,  
due to the need for extensive multi-view rendering and per-view inference, followed by complex geometric fusion. 
Finally, their performance relies heavily on LLM-driven prompt engineering on test sets, and they degrade significantly when presented with generic part labels common in real-world applications~\cite{garosi20253d}.

We propose \textbf{PatchAlign3D}, the first encoder-only 3D model that learns language-aligned \emph{local} features directly from point clouds. 
Our model bypasses the limitations of multi-view pipelines, achieving high-performance open-world 3D part segmentation in a single feed-forward pass.
We introduce a two-stage training strategy. Stage 1 performs 2D-to-3D feature distillation using a 3D transformer encoder, transferring dense visual features from a pre-trained 2D model such as DINOv2 \cite{oquab2023dinov2} to 3D patch tokens. This step lays the foundation for equipping 3D geometric encoding with fine-grained visual representations. Stage 2 then aligns these 3D patch embeddings with textual part descriptions encoded by a pre-trained text encoder, such as CLIP \cite{radford2021learning}.

A central challenge in our approach is learning from large-scale 3D shape segmentation data. We leverage 3D part annotations from Find3D \cite{ma2025find}, a recent data engine that automatically segments and annotates shape parts through a 2D segmentor SAM\cite{kirillov2023segment} and a VLM~\cite{team2023gemini}. These annotations are, however, inherently noisy and inconsistent across parts, for example, a single 3D patch may be associated with multiple part names, and segmentation masks are often fragmented or incomplete.
Key to our approach is that, instead of learning on unreliable point-level annotations, we perform semantic alignment at the \emph{patch level}. This local aggregation averages out annotation noise and is more robust to inconsistent boundaries. Furthermore, we propose a multi-positive sample-wise contrastive objective that uses fractional labels to handle ambiguous segmentation, yielding robust, generalizable geometric representations.

In summary, our contributions are: 
\begin{itemize}
    \item We introduce the first 3D encoder that produces language-aligned, patch-level features, closing the gap between global 3D foundation models and multi-view dense VLM pipelines.
    \item We propose a two-stage pre-training scheme operating at the patch level that effectively distills geometric representations from noisy, inconsistent part annotations via a multi-positive sample-wise contrastive objective.
    \item We demonstrate that PatchAlign3D achieves fast, single-pass zero-shot part segmentation, significantly outperforming both rendering-based and feed-forward baselines across multiple shape benchmarks.
\end{itemize}

\section{Related Work}
\label{sec:related}

\mypara{3D shape segmentation.}
Traditional 3D shape segmentation methods rely on fully supervised learning from part-annotated datasets such as ShapeNetPart~\cite{yi2016scalable}, PartNet~\cite{mo2019partnet}, and ScanObjectNN~\cite{uy2019revisiting}.
Early approaches focused on point-wise classification~\cite{Qi2016PointNetDL, Qi2017PointNetDH}, while later works explored prototype-based methods~\cite{He2020LearningAM, Yi2018GSPNGS} or co-segmentation objectives~\cite{Chen2019BAENETBA, Zhu2019AdaCoSegAS}.
Despite strong performance on known categories, these methods struggle to generalize to unseen objects and rely on costly manual supervision.
Scene-level segmentation methods~\cite{wu2024point} are more successful due to the amount of training data~\cite{Dai2017ScanNetR3, Yeshwanth2023ScanNetAH} and specialized networks~\cite{wu2024point} but cannot be directly applied to the more fine-grained domain of object-level part segmentation.
Our goal is to retain the feed-forward efficiency of supervised 3D models while enabling open-world part understanding.

\mypara{Adapting 2D foundation models to 3D.}
A parallel line of work explores the direct application of 2D foundation models, both uni-modal and language-vision models, to 3D by rendering point clouds from multiple views.
For scene segmentation, several works \cite{Nguyen2023Open3DISO3, 10655587, Peng2022OpenScene3S, Takmaz2023OpenMask3DO3} successfully extracted the 2D knowledge by projecting 2D features to 3D points. 
For shape analysis, PointCLIP~\cite{zhu2022pointclip} first showed that CLIP~\cite{radford2021learning} can perform zero-shot 3D classification from rendered images.
PointCLIPv2~\cite{zhu2022pointclip} further improves its performance and extends it to the more fine-grained task of part segmentation ,but with limited accuracy.
PartSLIP~\cite{liu2023partslip} adapted the GLIP~\cite{Li2021GroundedLP} bounding-box framework for object detection, while SATR~\cite{abdelreheem2023satr} refined this approach for mesh-based inputs.
Following the release of SAM~\cite{kirillov2023segment, ravi2024sam}, several works used it for class-agnostic 3D segmentation~\cite{Yang2024SAMPart3DSA, Liu2025PARTFIELDL3} with great success, and others combined it with 2D detectors for instance-level segmentation~\cite{Zhou2023PartSLIPEL, Xue2023ZeroPSHC}.
More recently, COPS~\cite{garosi20253d} used the powerful DINOv2~\cite{oquab2023dinov2} dense feature extractor to achieve state-of-the-art zero-shot shape segmentation results, though performance remains limited when using simple part text queries.
In contrast to these approaches, our method aims for a purely feed-forward 3D encoder that operates directly on point clouds.

\mypara{3D foundation models.}
Bridging the modality gap between 2D and 3D has led to the emergence of large-scale multimodal 3D foundation models.
ULIP~\cite{xue2023ulip} unified image, text, and 3D embeddings through cross-modal contrastive learning, while ULIP-2~\cite{xue2023ulip}, OpenShape~\cite{liu2023openshape}, and Uni3D~\cite{zhou2023uni3d} demonstrated that large-scale pre-training on Objaverse~\cite{deitke2023objaverse} enables strong zero-shot recognition.
However, these models primarily target global shape understanding rather than the more fine-grained part reasoning.
DITR~\cite{zeid2025dino} and OV3D~\cite{he2024unim} showed that distilling 2D features into a 3D encoder yields powerful geometric representations for scene segmentation,
while PartDistill~\cite{Umam2023PartDistill3S} attempted a similar strategy to one shape category at a time but with limited generalization across categories.
Find3D~\cite{ma2025find} advanced this direction by curating a 30K-shape Objaverse subset using SAM~\cite{kirillov2023segment} and Gemini~\cite{team2023gemini} captions to train an open-world feed-forward model.
Despite its scalability, Find3D still produces imprecise part boundaries and struggles with simple shapes.
Our work builds upon this data curation setup and introduces a two-stage pre-training framework that first transfers dense 2D features to 3D patches, then aligns them with language, achieving open-world, feed-forward part segmentation without any rendering at inference time.

\begin{figure*}[t]
  \centering
  \includegraphics[width=\linewidth]{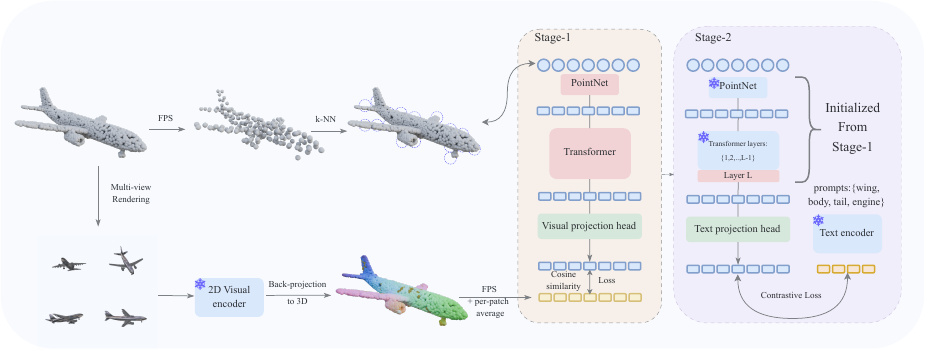}
    \caption{\textbf{PatchAlign3D pre-training.}
    Given an input point cloud, we extract multi-view visual features using a 2D backbone
    and back-project them into 3D space. In \textbf{Stage 1},
    the 3D transformer encoder operates on sampled point cloud patches and learns to
    align its output patch tokens with the back-projected visual features. In \textbf{Stage 2},
    we initialize from Stage 1, freeze all earlier layers, and train only the last
    transformer block and projector to align patch-level features with textual embeddings
    in a contrastive manner. For inference, we discard Stage 1 and use the pre-trained Stage 2 model.}
  \label{fig:arch}
\end{figure*}

\section{Method}
\label{sec:method}

\subsection{Overview}

We introduce \textbf{PatchAlign3D}, a two-stage framework for pre-training a point cloud transformer
to produce language-aligned local features and enable zero-shot 3D part segmentation.  
Following standard transformer-based 3D encoders such as PointBERT \cite{yu2022point}, our model operates on a sequence of patch tokens extracted from the input point cloud and outputs token features.

Our framework pre-trains a transformer encoder on these patch sets through two complementary stages, illustrated in \cref{fig:arch}:
\textbf{Stage 1} distills multi-view dense visual priors from a 2D feature extractor into a transformer-based point cloud encoder's token representation. This stage encourages the model to capture fine-grained rich features of strong 2D models, and serves as a robust initialization for the subsequent stage.
\textbf{Stage 2} aligns these 3D patch representations with textual embeddings in a contrastive approach to enable zero-shot local language reasoning. Importantly, operating at the patch level mitigates the noise and inconsistency present in Find3D's point-level annotations.
\textbf{At inference time}, we compare the embedding similarities between PatchAlign3D's output patch features and target textual queries, then the scores are propagated back to point-level labels.  
This two-stage approach produces strong features that enable downstream segmentation tasks without requiring multi-view rendering at inference time, relying solely on feed-forward shape processing.

\subsection{Data and Architecture}

\mypara{Training data.}
We follow the Find3D \cite{ma2025find} data engine's method to construct a 3D part annotation dataset. First, we select 32,052 shapes from a curated subset of Objaverse~\cite{deitke2023objaverse}, splitting them into 28,827 for training and 3,225 for validation. In total, we obtain more than 2 million part annotations across 761 object categories.
Each shape is rendered into 10 views, and multi-scale 2D masks for these views are generated using SAM~\cite{kirillov2023segment}.
Each masked view is then provided to the Gemini 1.5 \cite{team2023gemini} model, which predicts a single-word description for the masked region. These short labels (e.g., leg, wing, lid) are used as part-level text queries during both training and inference.
The 2D annotations are back-projected onto the corresponding 3D point cloud, producing per-point pseudo-labels.
Since points may appear in several rendered views, some receive multiple labels, which are retained as multi-label supervision during pre-training. 

\mypara{Architecture.}
Recent 3D foundation models \cite{xue2023ulip,liu2023openshape,zhou2023uni3d}
have demonstrated that transformer-based encoders, such as PointBERT \cite{yu2022point}, are effective for large-scale pre-training. We adopt a similar backbone.
The input point cloud is first partitioned into $G$ local patches $\{P_i\}_{i=1}^{G}$, each containing $k$ points. Patch centers ${c_i}$ are sampled using farthest-point sampling (FPS), and local neighborhoods are formed by selecting the $k$ nearest points around each centroid.

Each patch $P_i$ is encoded into a feature token through a lightweight PointNet. The corresponding centroid $c_i$ is embedded using a small MLP and added to the patch token as a positional encoding. The resulting sequence of $G$ tokens is then processed by a 12-layer vanilla transformer encoder~\cite{vaswani2017attention}.

\mypara{2D Dense Feature Extraction.}
In Stage 1, PatchAlign3D transfers semantic priors from 2D vision models into the 3D domain. To prepare these priors, we use a dense 2D feature extractor $\phi_{\text{2D}}$ that outputs a spatial feature field
$F_r(u,v)$ for each rendered view $r$.
We follow the multi-view feature extraction strategy of~\cite{garosi20253d} to project these features to the corresponding point cloud:
the 2D features are first upsampled to the original image resolution using bicubic interpolation and then back-projected to the 3D surface.
Each pixel $(u,v)$ corresponds to a visible surface point $x_{uv} \in \mathbb{R}^3$,
which inherits the 2D feature value $F_r(u,v)$ from that view.
Each 3D point aggregates features from all renderings in which it is visible:
\begin{equation}
    d(x) = \frac{1}{|\mathcal{V}(x)|}
\sum_{r \in \mathcal{V}(x)} F_r(u_r(x), v_r(x)),
\end{equation}
where $\mathcal{V}(x)$ denotes the set of views that observe point $x$.
Points not visible in any view are assigned interpolated features from nearby observed points using nearest-neighbor interpolation.
In practice, we use DINOv2~\cite{oquab2023dinov2} as $\phi_{\text{2D}}$ for all experiments,
ensuring a fair comparison with rendering-based baselines, though the procedure is model-agnostic
and compatible with any dense visual encoder.

To reduce storage and computation, we aggregate per-point features into patch-level
representations using the same procedure described earlier.
For each patch $P_i = \{x_m^{(i)}\}_{m=1}^{k}$, the target feature is computed as the mean
of its constituent point features:
\begin{equation}
d_i = \frac{1}{k} \sum_{m=1}^{k} d(x_m^{(i)}).
\end{equation}
For every shape, we cache the patch centers $c_i^{\text{cache}}$, memberships, and target features $\{d_i\}$.
This pre-processing produces consistent 3D features that serve as supervision for Stage 1 2D–3D feature distillation.

\subsection{Two-Stage  Pre-training}

\mypara{Stage 1: 2D-to-3D feature distillation.}
Each shape is partitioned into \(G\) patches of size \(k\) points, following the same above procedure.
To ensure correspondence with cached features, each online patch center $c_i^{\text{online}}$ is matched
to the nearest cached center $c_j^{\text{cache}}$.
The cached visual feature $d_{j}$ serves as the supervision for patch $P_i$.
The transformer encoder $f_\theta$ maps each patch to a latent token
$z_i = f_\theta(P_i) \in \mathbb{R}^d$,
which is projected to the 2D feature space through a linear head $h_{\text{2D}}$.
The model is optimized with a cosine-similarity regression loss:
\begin{equation}
\mathcal{L}_{\text{2D}} =
\frac{1}{G} \sum_{i=1}^{G}
\Big[ 1 -
\frac{h_{\text{2D}}(z_i)^\top d_i}{
\|h_{\text{2D}}(z_i)\|_2 \, \|d_i\|_2}
\Big].
\end{equation}
This pre-training step transfers rich fine-grained representations from the 2D feature space to the 3D encoder in a self-supervised manner, and prepares the encoder for the coarser, language-based supervision of Stage 2.

\mypara{Stage 2: 3D–text patch contrastive learning.}
We initialize Stage 2 from the Stage 1 checkpoint and freeze the early layers to preserve the geometry-aware representations and prevent catastrophic forgetting. Only the final transformer block and a lightweight linear head $h_{\text{text}}$ are trained to align patch embeddings with the text encoder's feature space.
To supervise this alignment, we adopt the sigmoid-based contrastive formulation introduced in SigLIP~\cite{zhai2023sigmoid}, which replaces the standard softmax normalization. This design is shown to be more effective with a small number of negatives.
For each shape, we consider $G$ patches indexed by $i \in \{1,\ldots,G\}$ and $C_s$ valid part categories indexed by $j \in \{1,\ldots,C_s\}$. Let $t_j \in \mathbb{R}^d$ denote the text embedding corresponding to part $j$, obtained from a pretrained text encoder. Patch–text similarity scores are computed as
\begin{equation}
    s_{i,j} = \frac{1}{\tau} \left\langle h_{\text{text}}(z_i),\, t_j \right\rangle + b,
\end{equation}
where the temperature $\tau > 0$ and bias $b$ are learnable parameters, initialized to $0.1$ and $-10$, respectively. The similarities are converted to probabilities through a sigmoid activation $\sigma(s_{i,j})$.

To add robustness to uncertain boundaries and noisy ground-truth segmentation masks, each patch–part pair is assigned a fractional label $y_{i,j} \in [0,1]$ proportional to the fraction of points in patch $P_i$ that belong to part $j$. Hence, $y_{i,j} = 1$ when all points belong to part $j$, and $y_{i,j} > 0$ when the patch partially overlaps that part. Each patch can be associated with multiple text annotations, which leads to a multi-positive contrastive objective.

Negatives are defined \emph{within each sample} as entries with $y_{i,j} = 0$, rather than across the batch. Using only within-sample negatives avoids treating identical parts in different shapes as negatives, which would otherwise harm open-world generalization. 

Training minimizes a sigmoid binary cross-entropy loss:
\begin{equation}
\scalebox{0.75}{$
\mathcal{L}_{\text{text}} =
\sum_{i=1}^{G} \sum_{j=1}^{C_s}
\big[
 -\, y_{i,j} \log \sigma(s_{i,j})
 - (1 - y_{i,j}) \log (1 - \sigma(s_{i,j}))
\big].
$}
\end{equation}
To further improve generalization, we apply geometric data augmentations during training, including random rotations, translations, scaling, and point jittering. 

\mypara{Inference.}
At test time, for zero-shot segmentation, we compute the similarity between the 3D encoder's output patch features and the text features of the target part names. Each patch is assigned the label corresponding to the maximum similarity score, and patch predictions are upsampled to point-level labels by assigning each point the label of its nearest patch centroid.

\newcommand{\rowlabel}[1]{%
  \smash{\raisebox{30pt}{\rotatebox[origin=c]{90}{\scriptsize #1}}}%
}

\begin{figure*}[tp]
    \centering
    \setlength{\tabcolsep}{1pt}
    \renewcommand{\arraystretch}{1} 

    \begin{tabular}{@{}c*{6}{c}@{}}
        \rowlabel{Ground truth} &
        \includegraphics[width=0.14\textwidth]{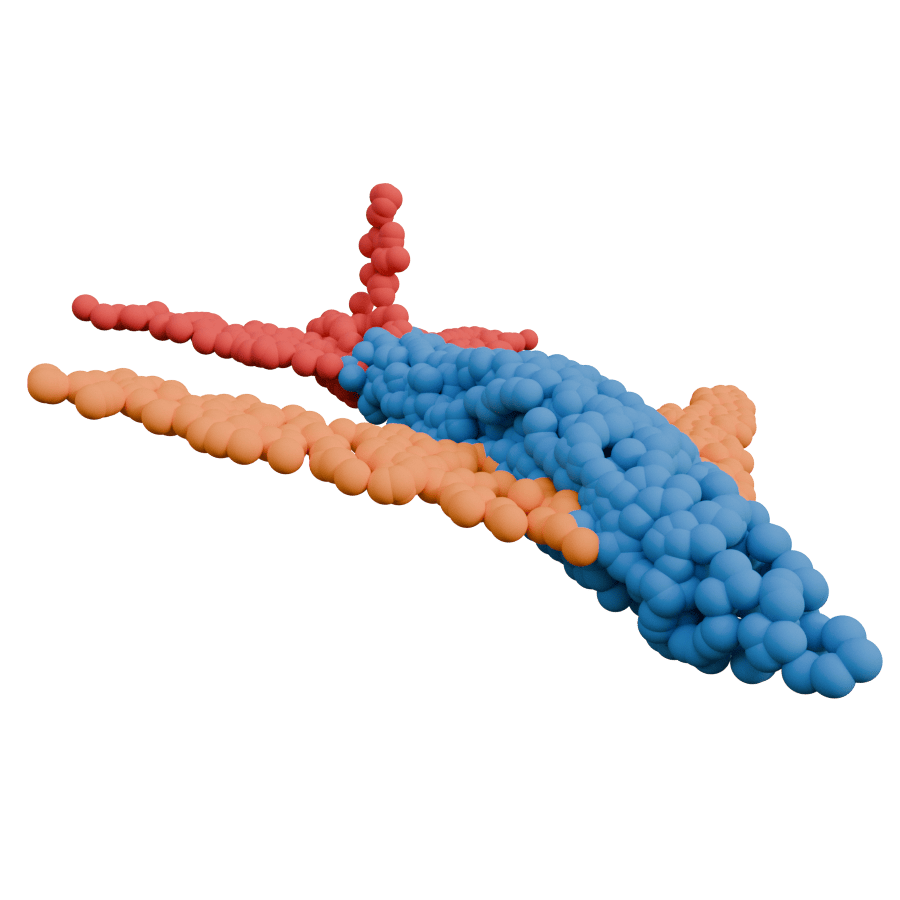} &
        \includegraphics[width=0.14\textwidth]{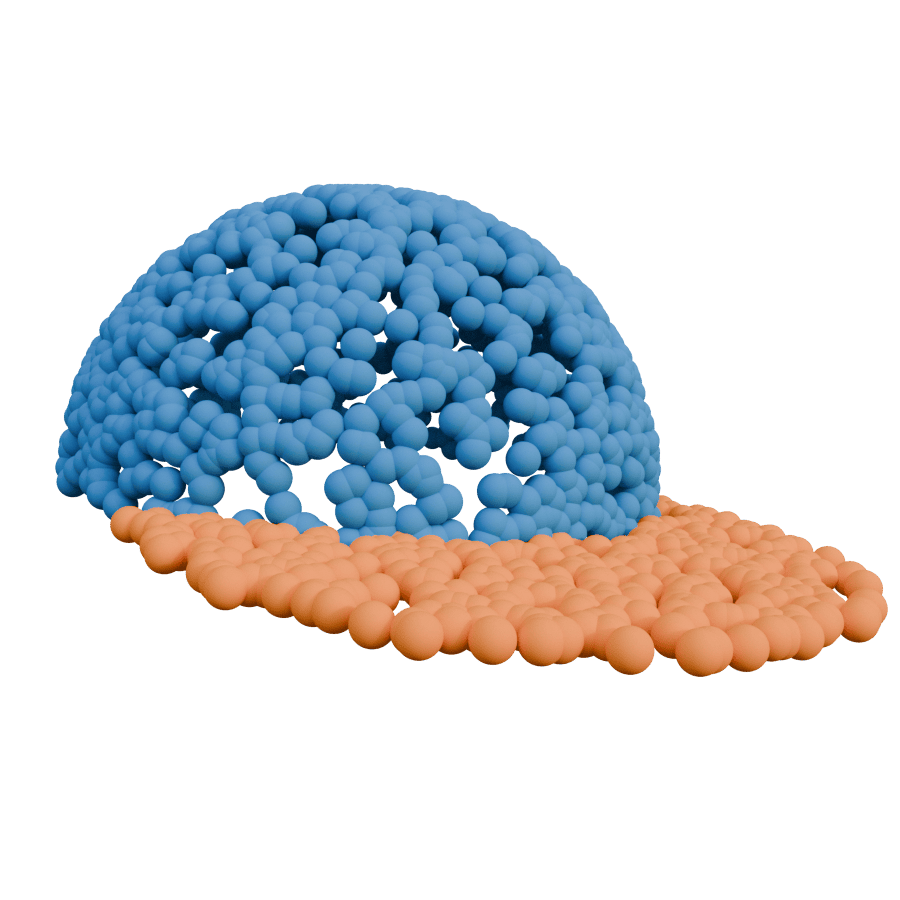} &
        \includegraphics[width=0.14\textwidth]{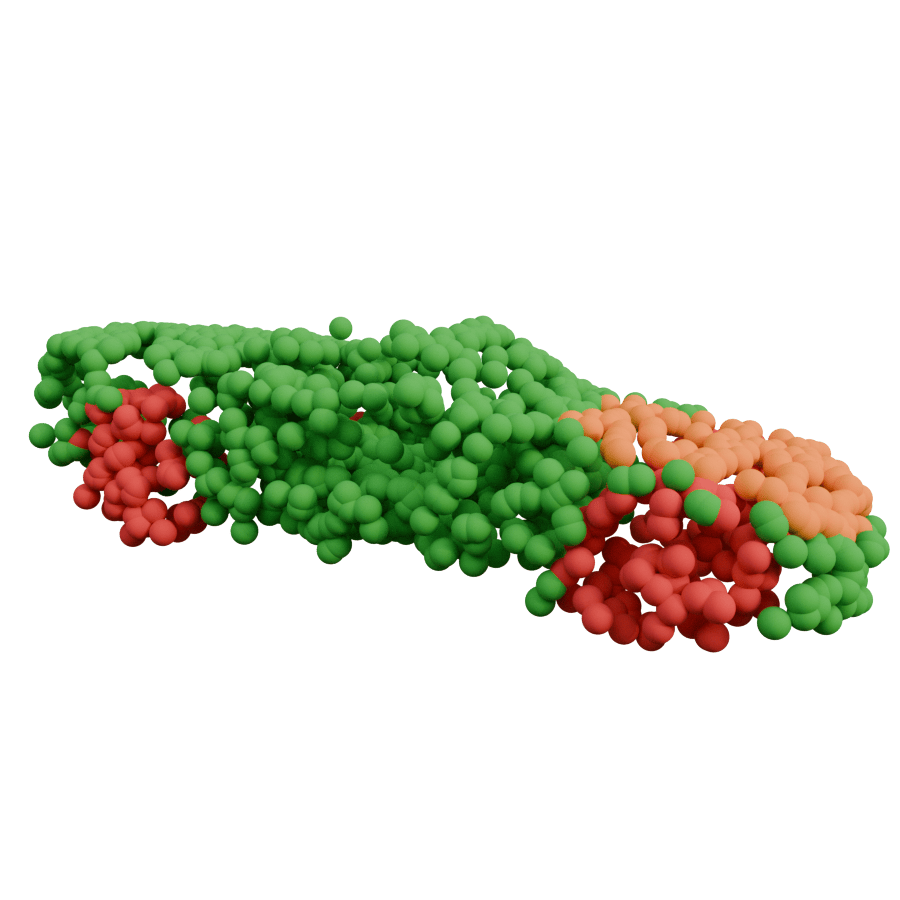} &
        \includegraphics[width=0.14\textwidth]{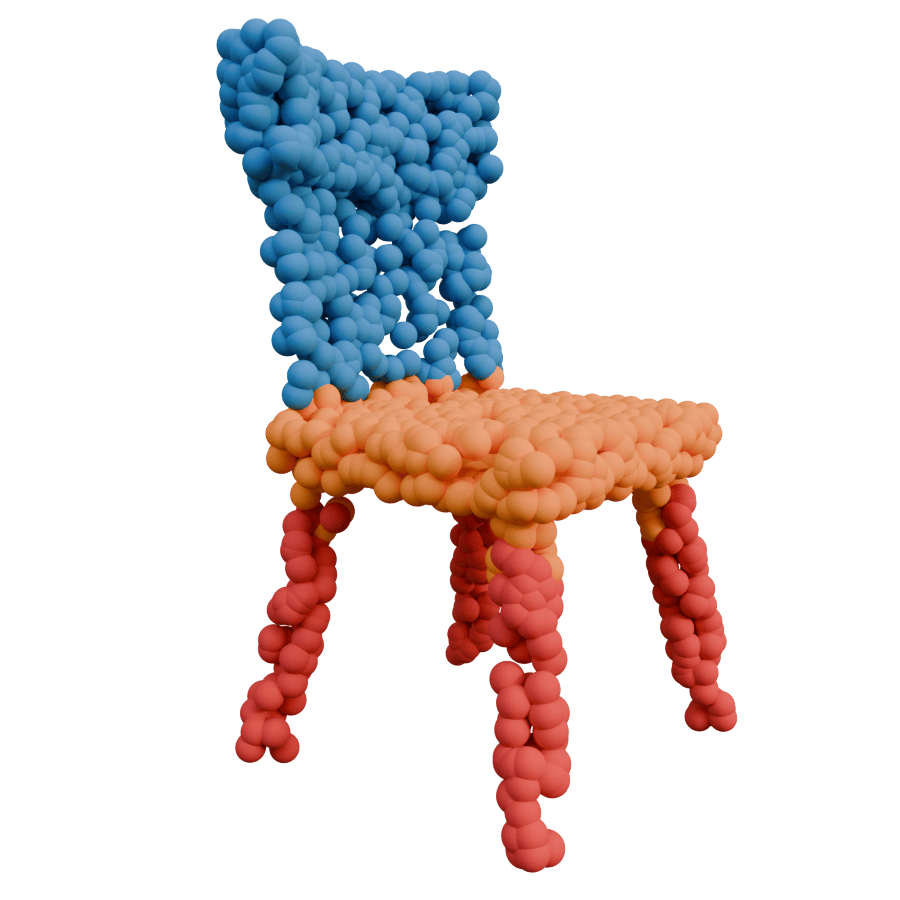} &
        \raisebox{-0.12\height}{\includegraphics[width=0.14\textwidth]{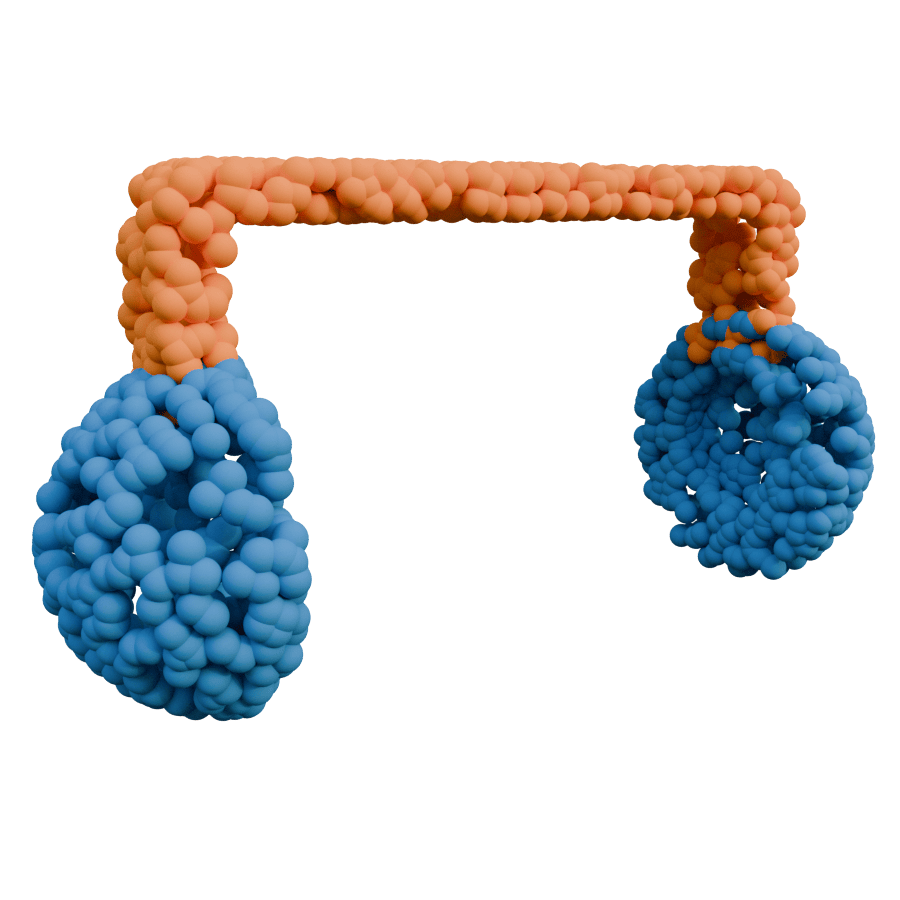}} &
        \raisebox{-0.12\height}{\includegraphics[width=0.14\textwidth]{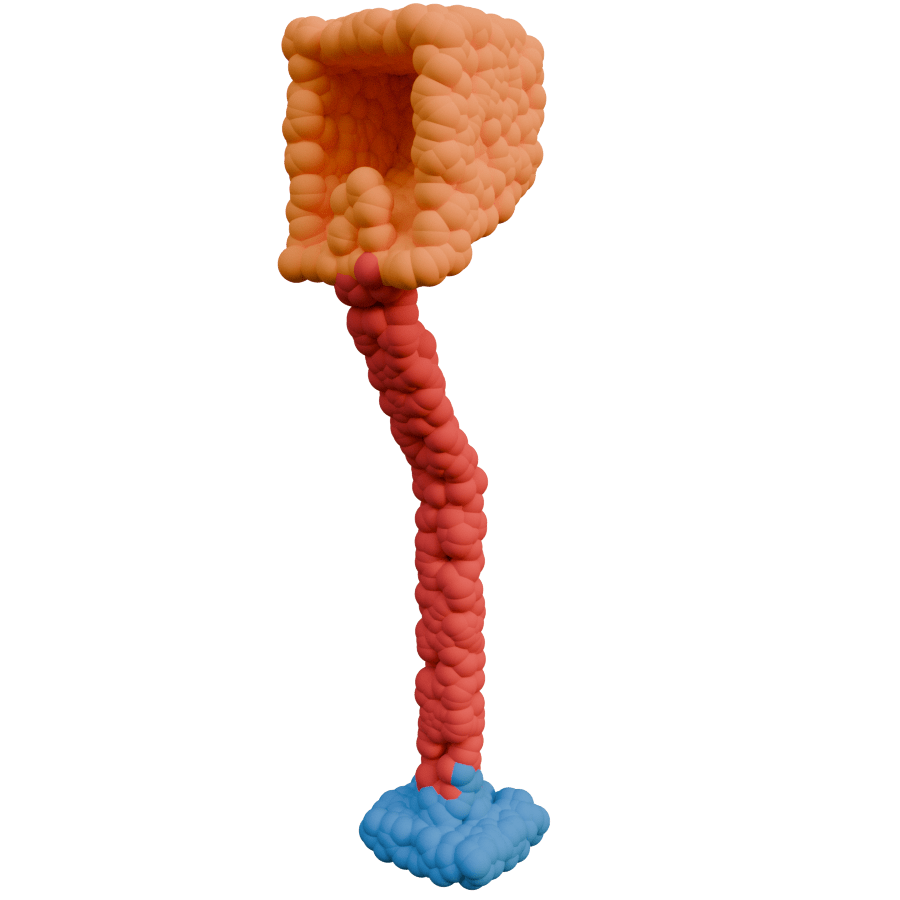}} \\[-10pt]

        \rowlabel{COPS} &
        \includegraphics[width=0.14\textwidth]{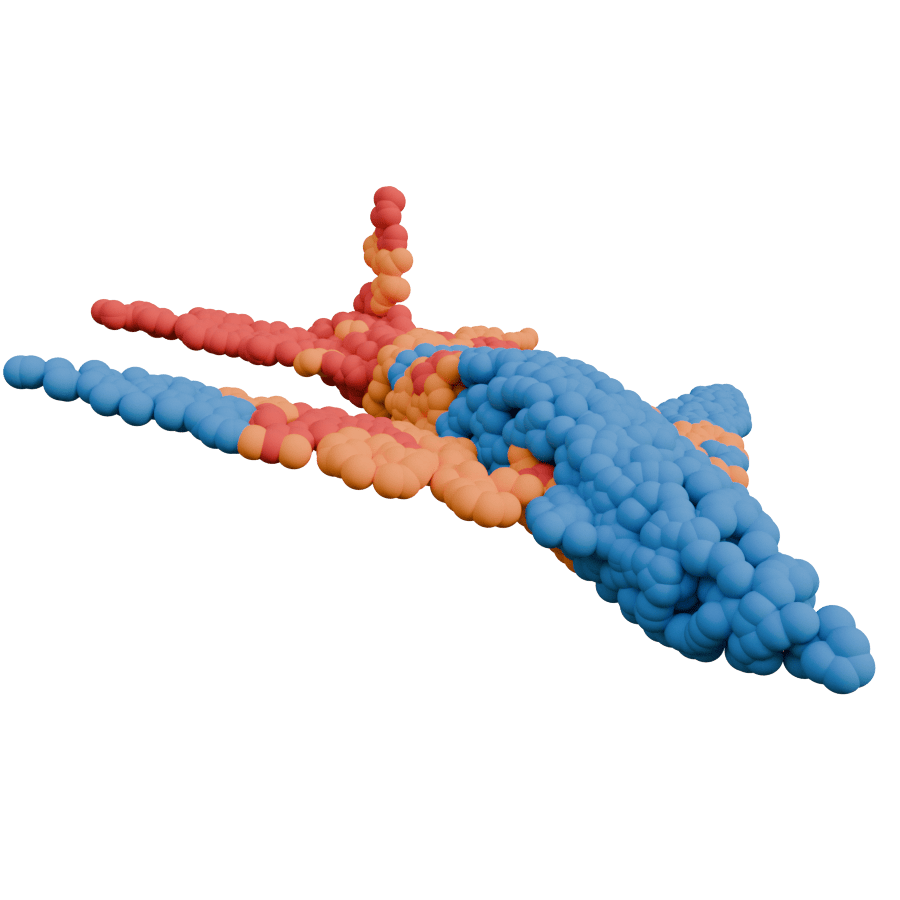} &
        \includegraphics[width=0.14\textwidth]{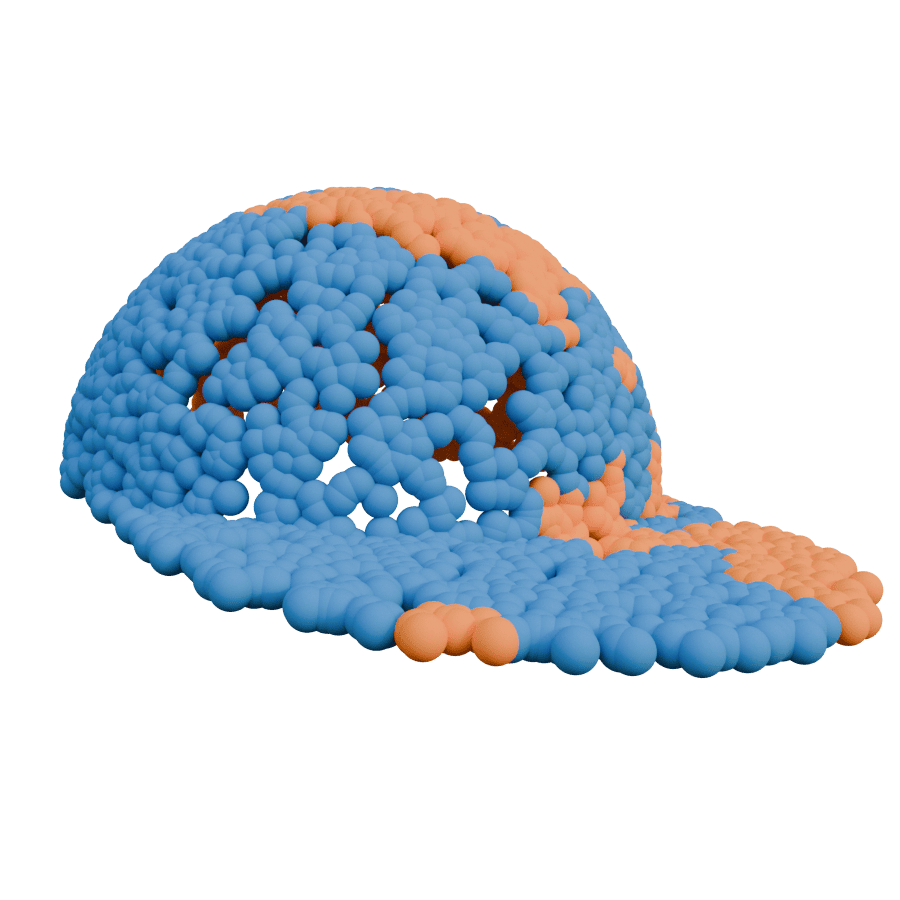} &
        \includegraphics[width=0.14\textwidth]{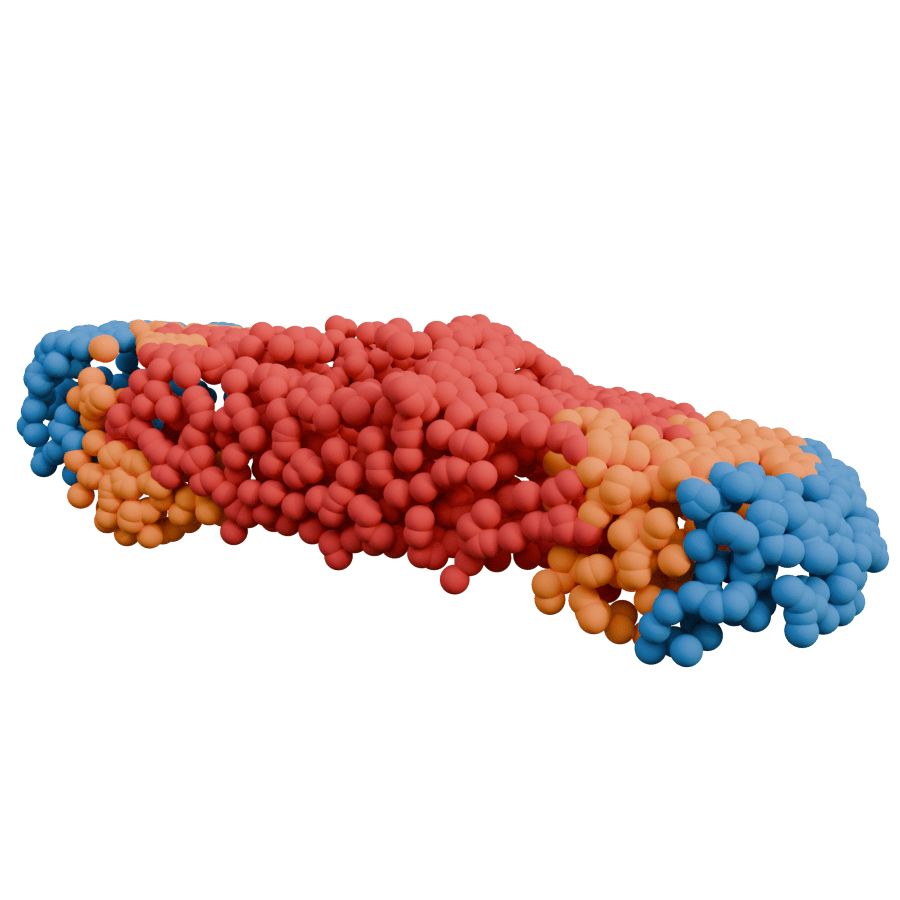} &
        \includegraphics[width=0.14\textwidth]{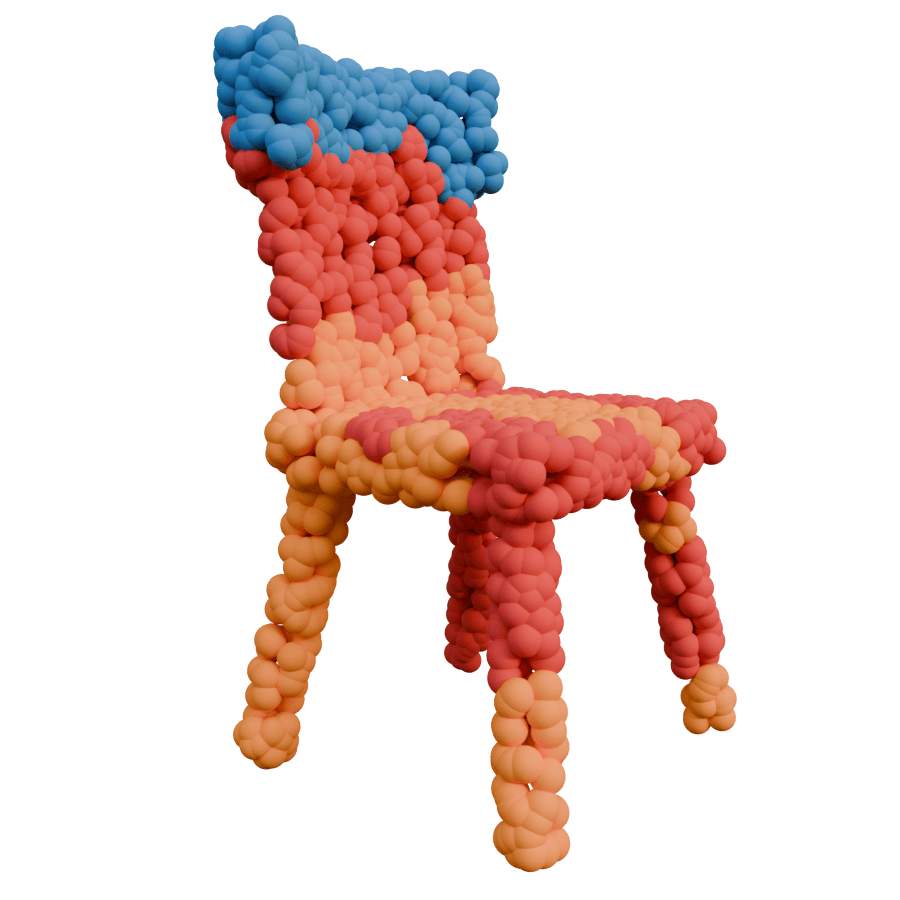} &
        \raisebox{-0.12\height}{\includegraphics[width=0.14\textwidth]{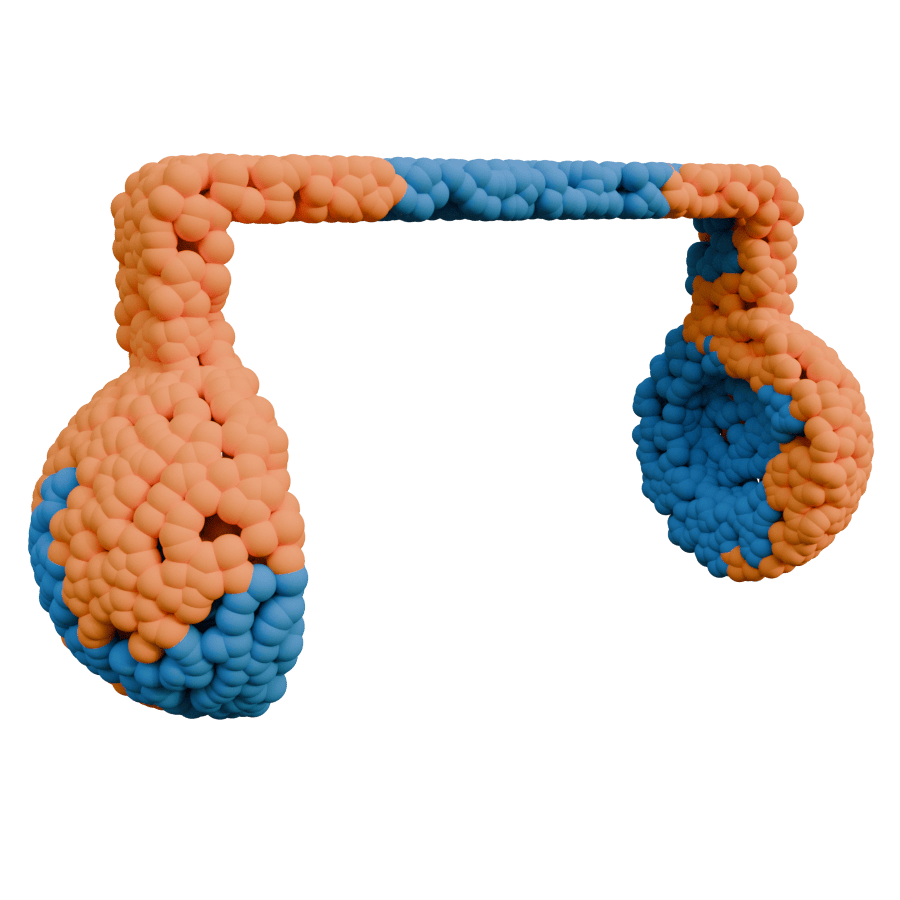}} &
        \raisebox{-0.12\height}{\includegraphics[width=0.14\textwidth]{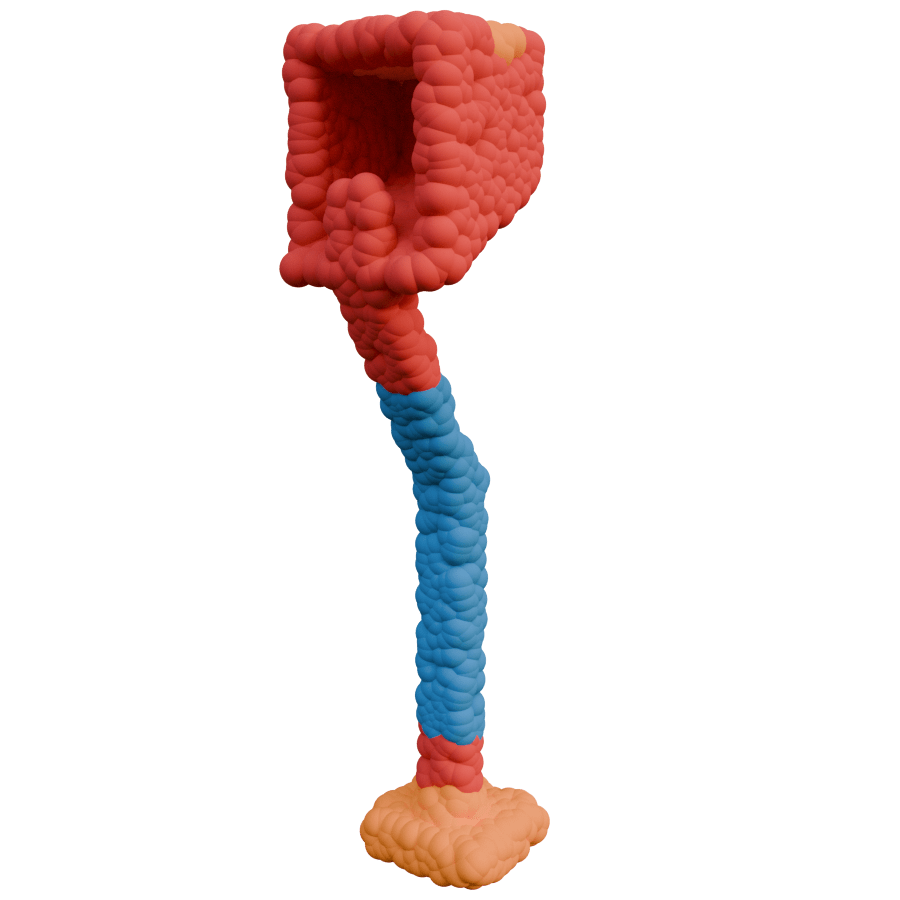}} \\[-10pt]

        \rowlabel{Find3D} &
        \includegraphics[width=0.14\textwidth]{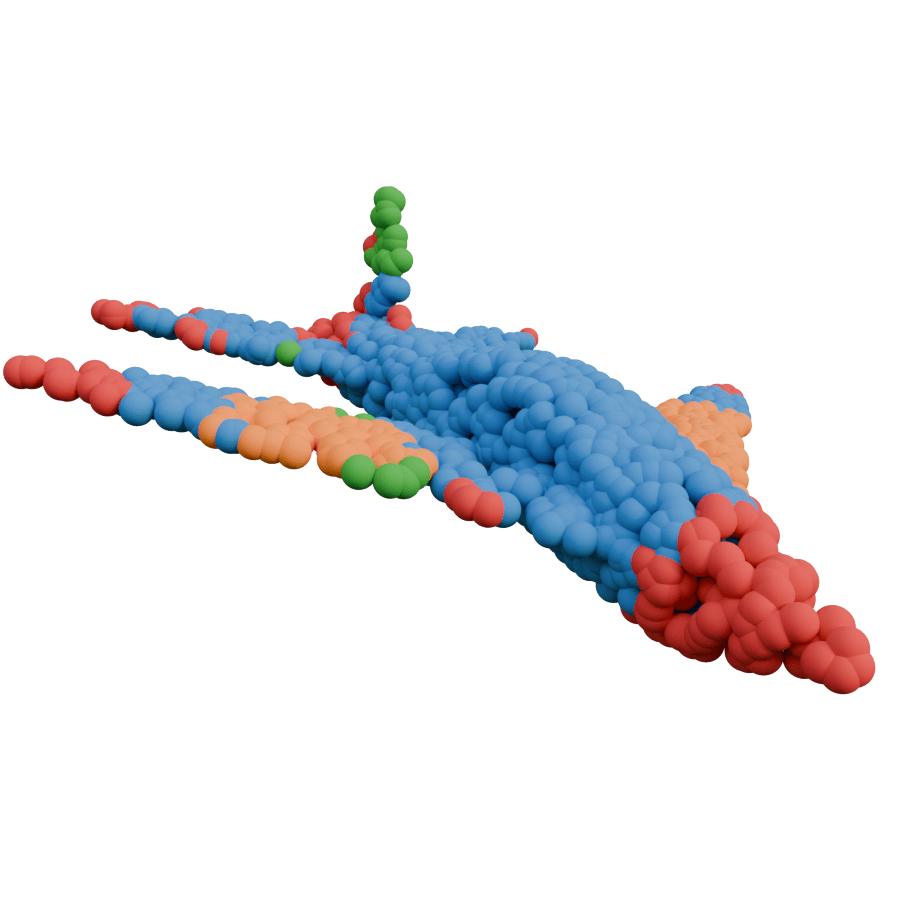} &
        \includegraphics[width=0.14\textwidth]{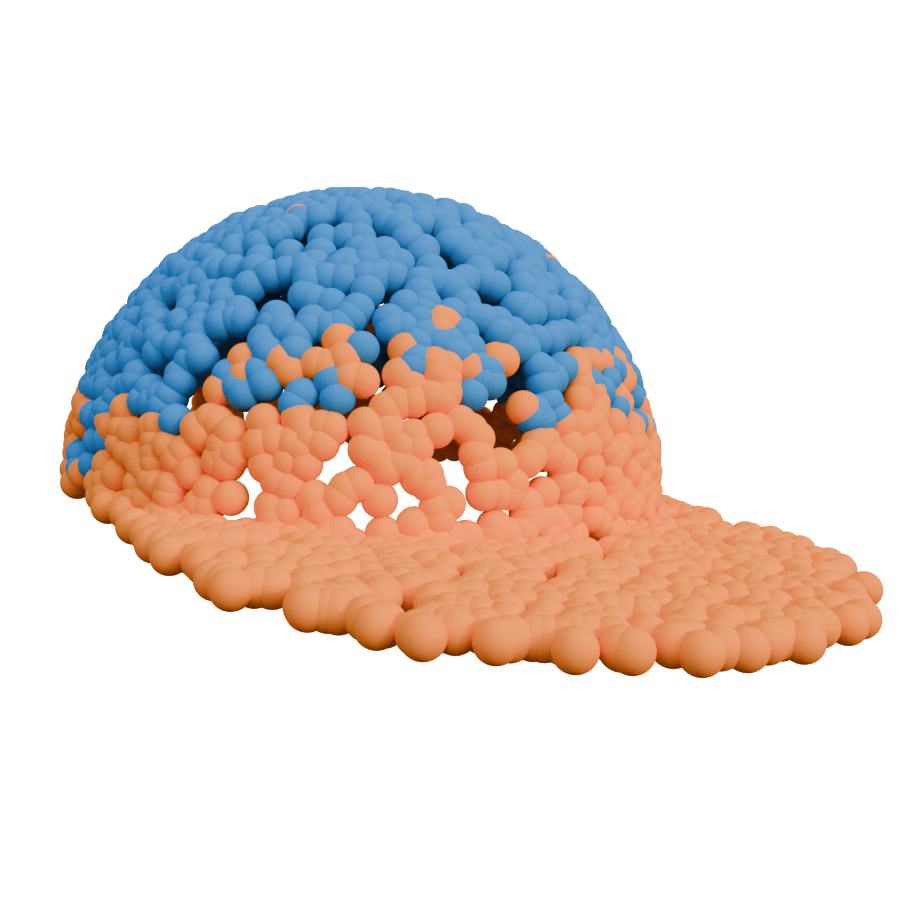} &
        \includegraphics[width=0.14\textwidth]{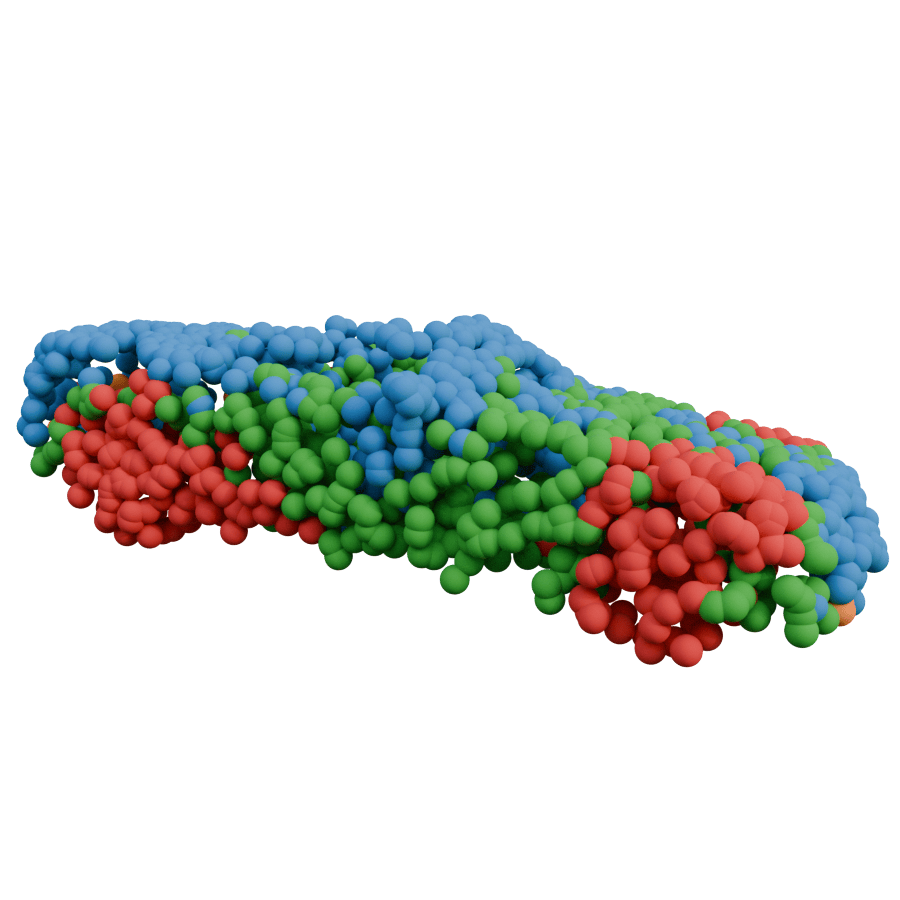} &
        \includegraphics[width=0.14\textwidth]{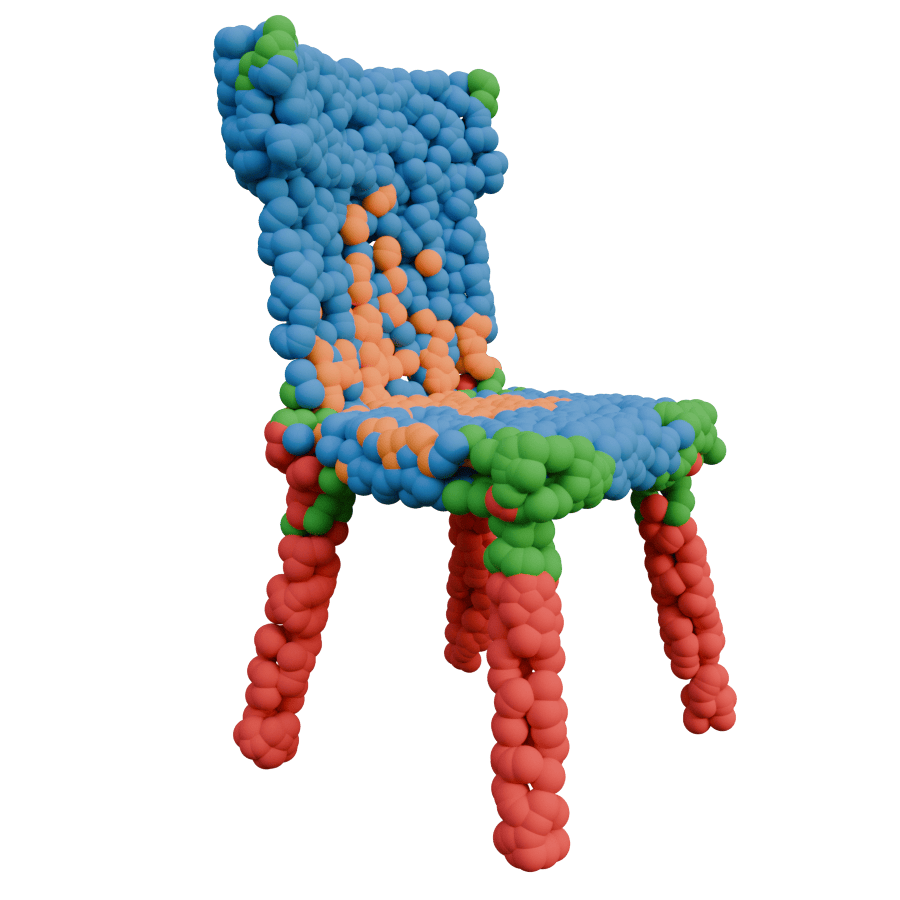} &
        \raisebox{-0.12\height}{\includegraphics[width=0.14\textwidth]{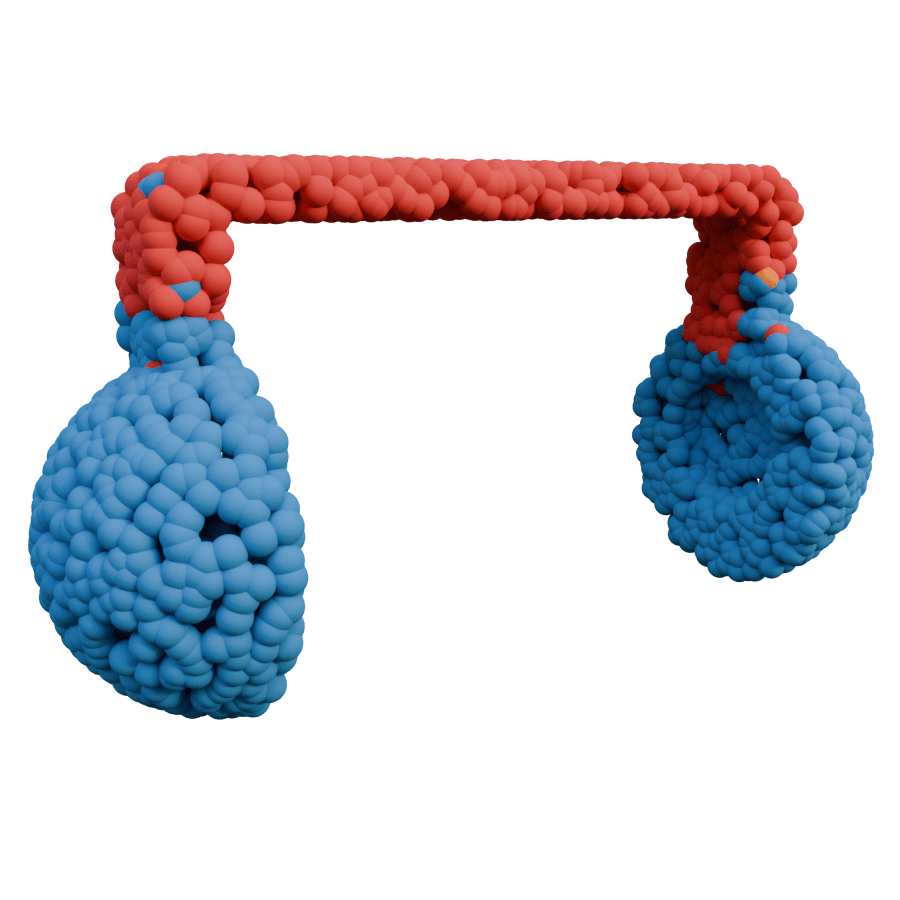}} &
        \raisebox{-0.12\height}{\includegraphics[width=0.14\textwidth]{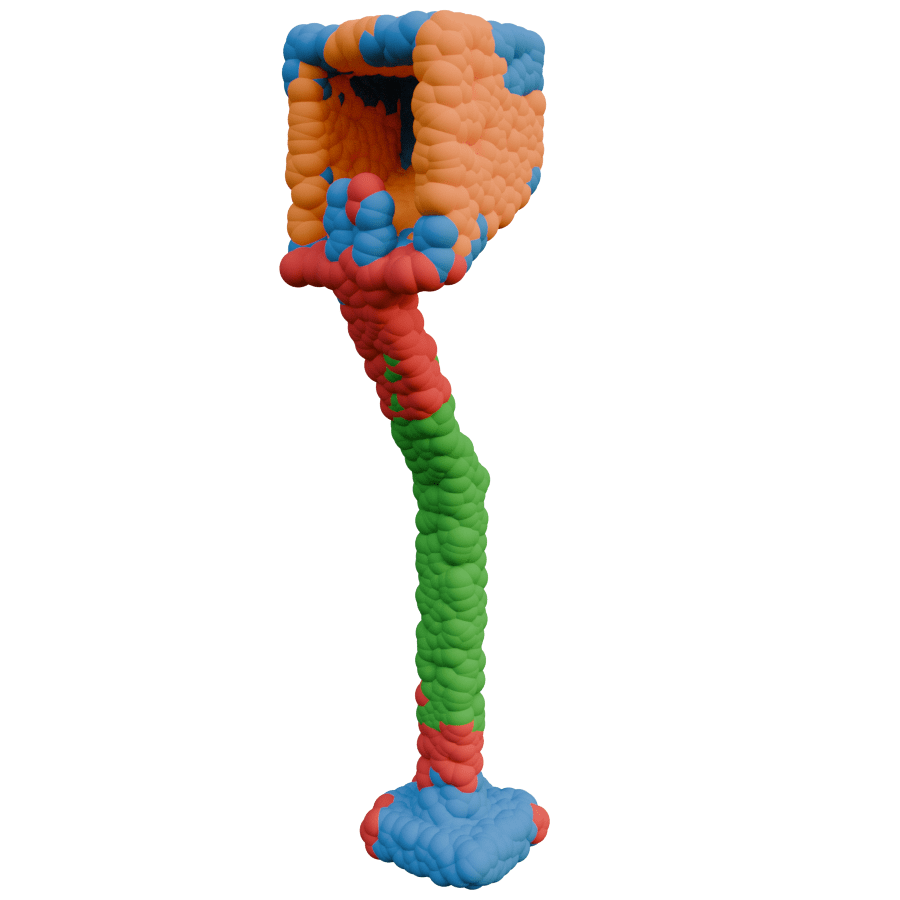}} \\[-10pt]

        \rowlabel{PatchAlign3D} &
        \includegraphics[width=0.14\textwidth]{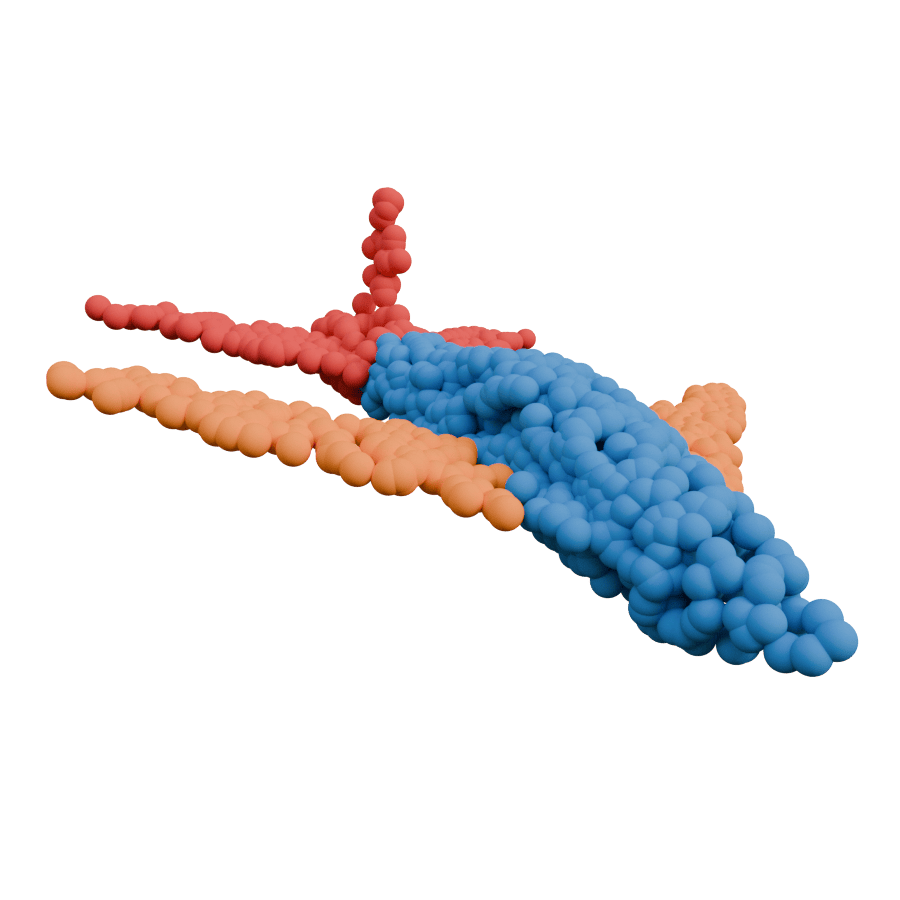} &
        \includegraphics[width=0.14\textwidth]{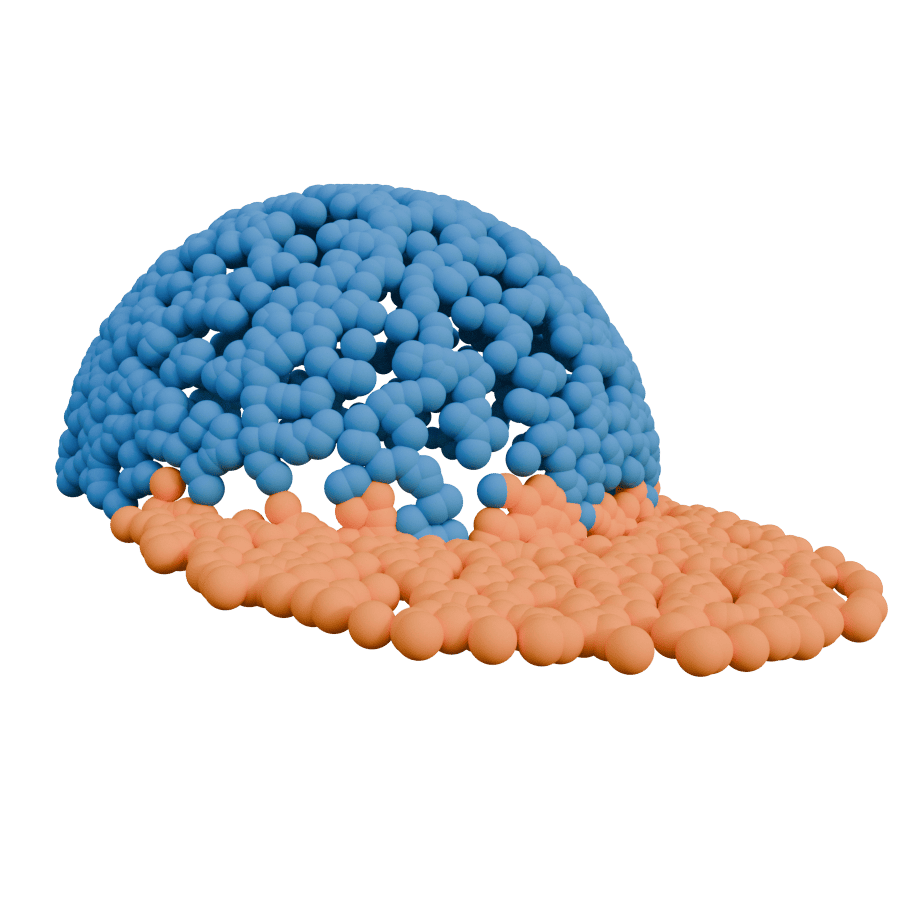} &
        \includegraphics[width=0.14\textwidth]{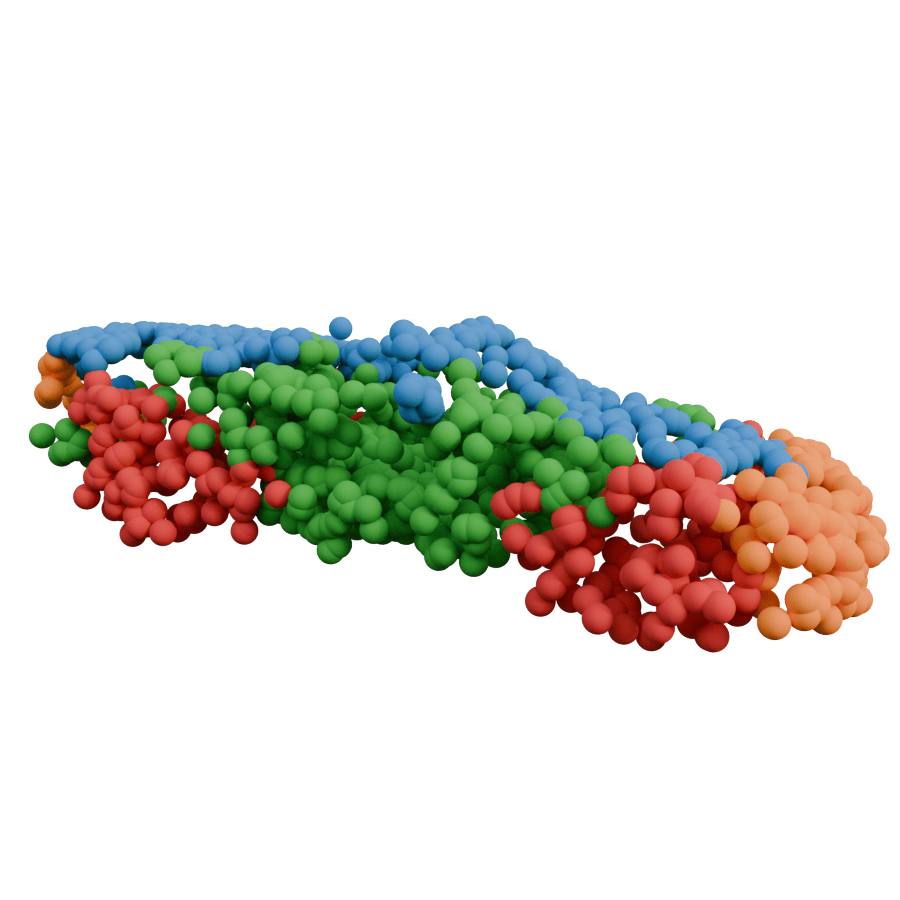} &
        \includegraphics[width=0.14\textwidth]{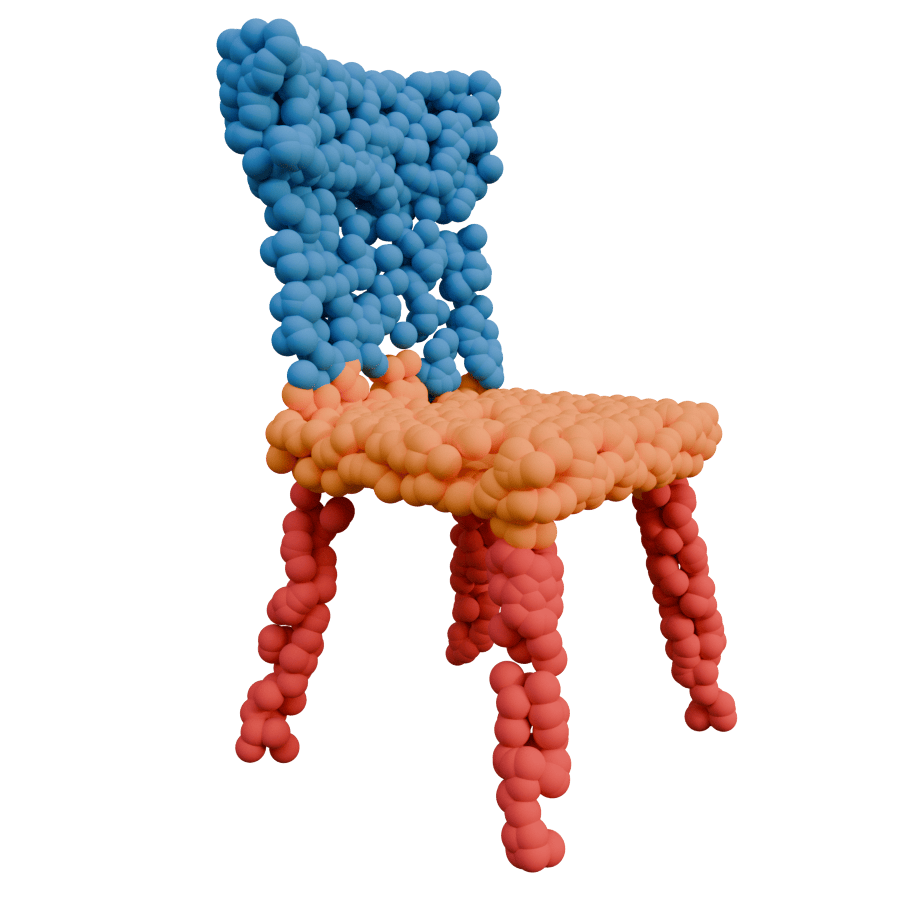} &
        \raisebox{-0.12\height}{\includegraphics[width=0.14\textwidth]{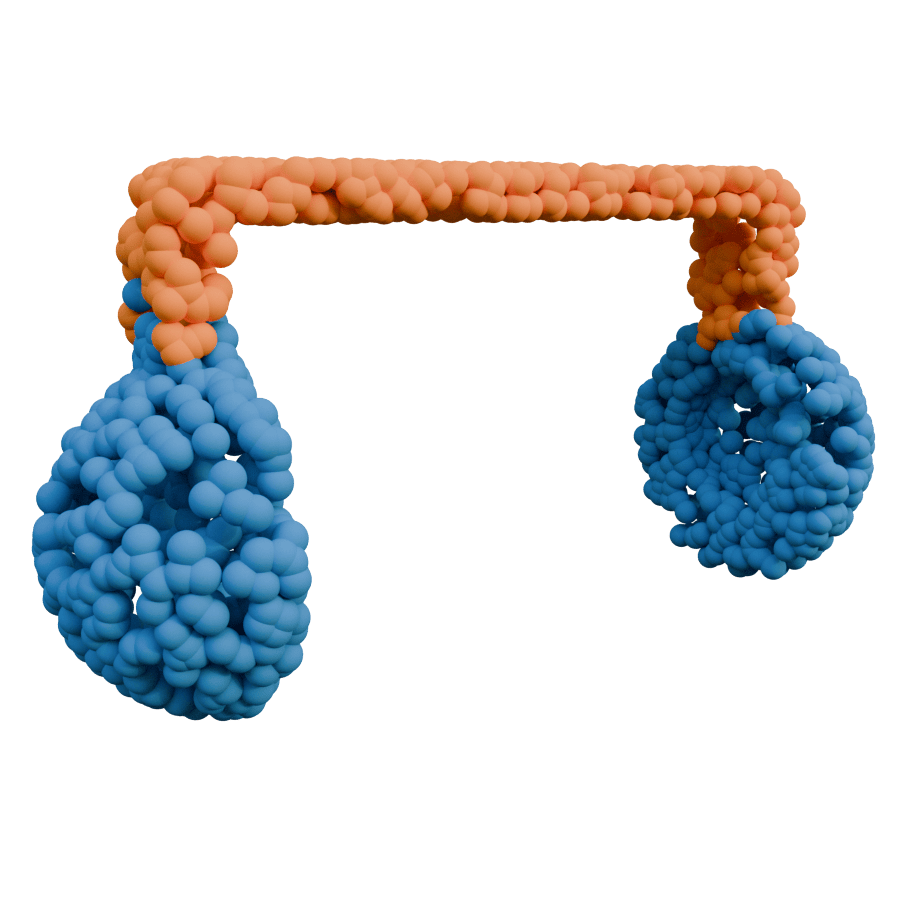}} &
        \raisebox{-0.12\height}{\includegraphics[width=0.14\textwidth]{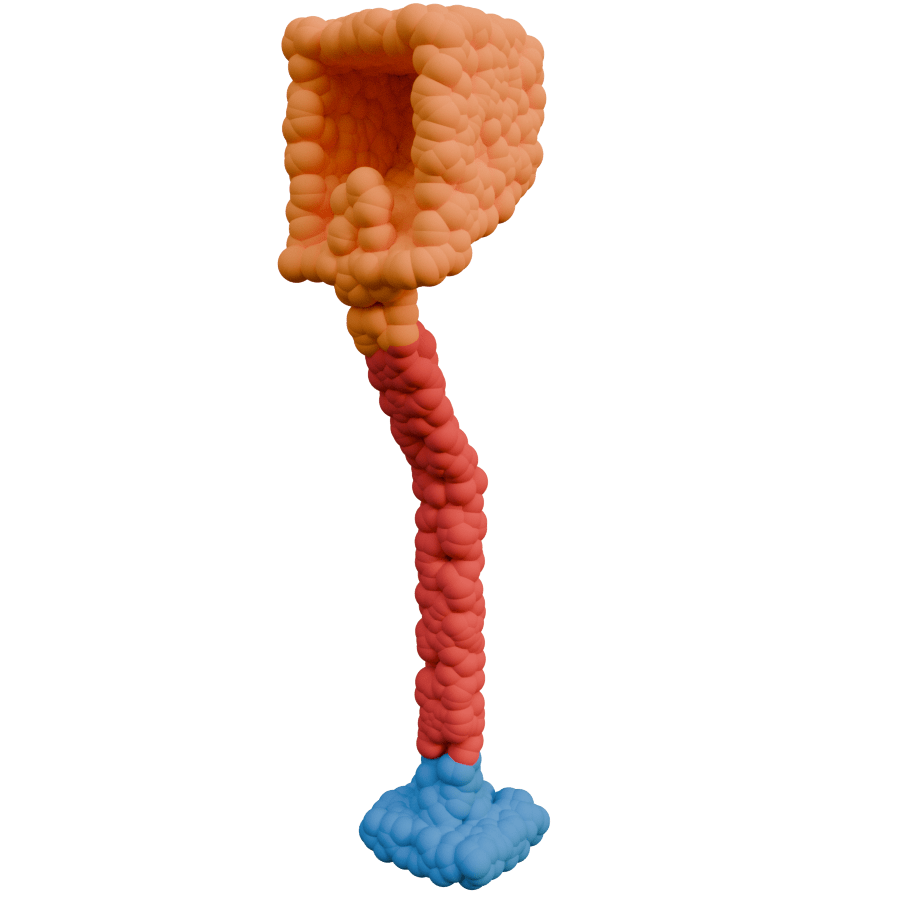}} \\[0pt]

        & {\scriptsize \{\textcolor{blue}{body}, \textcolor{orange}{wing}, \textcolor{red}{tail}, \textcolor{darkgreen}{engine}\}} &
          {\scriptsize \{\textcolor{blue}{crown}, \textcolor{orange}{brim}\}} &
          {\scriptsize \{\textcolor{blue}{roof}, \textcolor{orange}{hood}, \textcolor{red}{wheel}, \textcolor{darkgreen}{body}\}} &
          {\scriptsize \{\textcolor{blue}{back}, \textcolor{orange}{seat}, \textcolor{red}{leg}, \textcolor{darkgreen}{arm}\}} &
          {\scriptsize \{\textcolor{blue}{earcup}, \textcolor{orange}{headband}, \textcolor{red}{wire}\}} &
          {\scriptsize \{\textcolor{blue}{base}, \textcolor{orange}{shade}, \textcolor{red}{bracket}, \textcolor{darkgreen}{pole}\}}
    \end{tabular}

    \caption{\textbf{Qualitative comparisons on ShapeNetPart \cite{yi2016scalable}.} 
    We show ground truth (top row) and predictions from COPS~\cite{garosi20253d}, Find3D~\cite{ma2025find}, and PatchAlign3D 
    (rows 2–4) across six representative shapes. 
    The part legends below each column indicate the semantic labels used for zero-shot prediction. 
    PatchAlign3D produces noticeably more precise and coherent segmentations, despite relying 
    solely on an encoder and patch-level features.}
    \label{fig:qual_4x6}
\end{figure*}

\section{Experiments}
\label{sec:experiments}

We evaluate our model on a diverse set of zero-shot part segmentation benchmarks that include both synthetic and real-world inputs.  
Across all datasets, our approach consistently achieves state-of-the-art accuracy while maintaining a simple feed-forward architecture and high inference speed.

\subsection{Benchmarks}

\mypara{Datasets.}
We evaluate PatchAlign3D on five zero-shot part segmentation benchmarks covering synthetic, scanned, rigid, and non-rigid shapes: ShapeNetPart~\cite{yi2016scalable}, PartNetE~\cite{liu2023partslip}, ScanObjectNN~\cite{uy2019revisiting}, FAUST~\cite{bogo2014faust}, and Objaverse–General \cite{ma2025find, deitke2023objaverse}. 
ShapeNetPart~\cite{yi2016scalable} contains 2,874 test point clouds across 16 categories with fine-grained part labels and serves as the standard benchmark for zero-shot segmentation of human-made objects. 
PartNetE~\cite{liu2023partslip} provides a curated subset of 1,906 shapes with detailed annotations, though not all points are labeled. 
FAUST~\cite{bogo2014faust} consists of 300 non-rigid human body scans, covering multiple poses. For the segmentation benchmark, we use the coarse part annotations proposed by SATR~\cite{abdelreheem2023satr}.
ScanObjectNN includes 2,902 real objects captured from cluttered, noisy scenes. 
For Objaverse–General, we follow Find3D \cite{ma2025find} and select 100 shapes from the validation split of the training data. Since the original seen/unseen split is not provided, we propose a split with 14 unseen categories relative to the training data. 
Together, these datasets cover a broad spectrum of shape categories and types, levels of realism, and geometric variability.

\begin{table*}[t]
\centering
\resizebox{\linewidth}{!}{%
\begin{tabular}{c|l|cc|cccccccccccccccc}
\hline
\textcolor{gray}{Pipeline} & Method & mIoU & cIoU & Airplane & Bag & Cap & Car & Chair & Earph. & Guitar & Knife & Lamp & Laptop & Motor. & Mug & Pistol & Rocket & Skate & Table \\ \hline
\multicolumn{20}{c}{\textbf{Mesh methods}} \\ \hline
\textcolor{gray}{Rendering} & 3DH~\cite{decatur20233d}  & 9.6 & 5.7 & 5.8 & 2.1 & 2.9 & 2.9 & 15.5 & 9.6 & 0.9 & 1.6 & 13.2 & 1.8 & 5.6 & 0.7 & 1.4 & 10.4 & 6.4 & 10.8 \\
\textcolor{gray}{Rendering} & SATR~\cite{abdelreheem2023satr} & 32.8 & 31.9 & 38.5 & 44.6 & 24.0 & 19.6 & 33.2 & 16.9 & 40.2 & 45.9 & 30.2 & 37.8 & 15.7 & 52.3 & 20.9 & 28.4 & 30.8 & 31.4 \\ \hline
\multicolumn{20}{c}{\textbf{Point-cloud methods}} \\ \hline
\textcolor{gray}{Rendering}   & PointCLIPv2~\cite{zhu2022pointclip} & 16.1 & 21.0 & 5.98 & 16.4 & 34.5 & 17.1 & 15.8 & 41.6 & 19.9 & 45.4 & 36.5 & 30.6 & 2.5 & 24.7 & 22.6 & 10.4 & 16.2 & 10.9 \\
\textcolor{gray}{Rendering}   & COPS~\cite{garosi20253d}     & 25.6 & 32.2 & 13.8 & 31.0 & 46.1 & 10.4 & 23.2 & 44.2 & 40.2 & 60.1 & 42.1 & \textbf{63.3} & 7.6 & 39.0 & 32.3 & 17.2 & 25.8 & 19.6 \\
\textcolor{gray}{Feed-forward} & Find3D~\cite{ma2025find}     & 23.3 & 23.9 & 15.6 & 10.7 & 13.9 & 13.2 & 27.2 & 50.9 & 22.7 & 31.5 & 27.1 & 31.1 & 14.1 & 10.5 & 13.5 & 26.4 & 51.9 & 22.8 \\
\rowcolor{OursTint}
\textcolor{gray}{Feed-forward} & PatchAlign3D & \textbf{56.9} & \textbf{53.1} & \textbf{51.7} & \textbf{51.1} & \textbf{65.8} & \textbf{32.5} & \textbf{65.3} & \textbf{66.7} & \textbf{49.5} & \textbf{63.6} & \textbf{52.0} & 61.1 & \textbf{23.0} & \textbf{69.3} & \textbf{37.1} & \textbf{37.7} & \textbf{62.3} & \textbf{60.8} \\ \hline
\multicolumn{2}{c|}{Improvement} 
  & \textcolor{PosGreen}{\bf +31.3} & \textcolor{PosGreen}{\bf +20.9}
  & \textcolor{PosGreen}{\bf +36.1} & \textcolor{PosGreen}{\bf +20.1} & \textcolor{PosGreen}{\bf +19.7} & \textcolor{PosGreen}{\bf +15.4}
  & \textcolor{PosGreen}{\bf +38.1} & \textcolor{PosGreen}{\bf +15.8} & \textcolor{PosGreen}{\bf +9.3} & \textcolor{PosGreen}{\bf +3.5}
  & \textcolor{PosGreen}{\bf +9.9} & \textcolor{NegRed}{\bf -2.2} & \textcolor{PosGreen}{\bf +8.9}
  & \textcolor{PosGreen}{\bf +30.3} & \textcolor{PosGreen}{\bf +4.8} & \textcolor{PosGreen}{\bf +11.3}
  & \textcolor{PosGreen}{\bf +10.4} & \textcolor{PosGreen}{\bf +38.0} \\ \hline
\end{tabular}%
}
\caption{\textbf{Zero-shot part segmentation results on ShapeNetPart \cite{yi2016scalable}.} We directly compare PatchAlign3D with state-of-the-art point cloud methods. Mesh-based approaches are included for reference. PatchAlign3D significantly outperforms the strongest baselines.}
\label{tab:shapenetpart}
\end{table*}

\newcommand{\rowlabelnew}[1]{%
  \smash{\raisebox{30pt}{\rotatebox[origin=c]{90}{\scriptsize #1}}}%
}
\begin{figure}[t]
    \centering
    \setlength{\tabcolsep}{1pt}
    \renewcommand{\arraystretch}{0}

\begin{tabular}{@{}c*{3}{c}@{}}
    \rowlabelnew{Ground truth} &
    \includegraphics[width=0.30\linewidth]{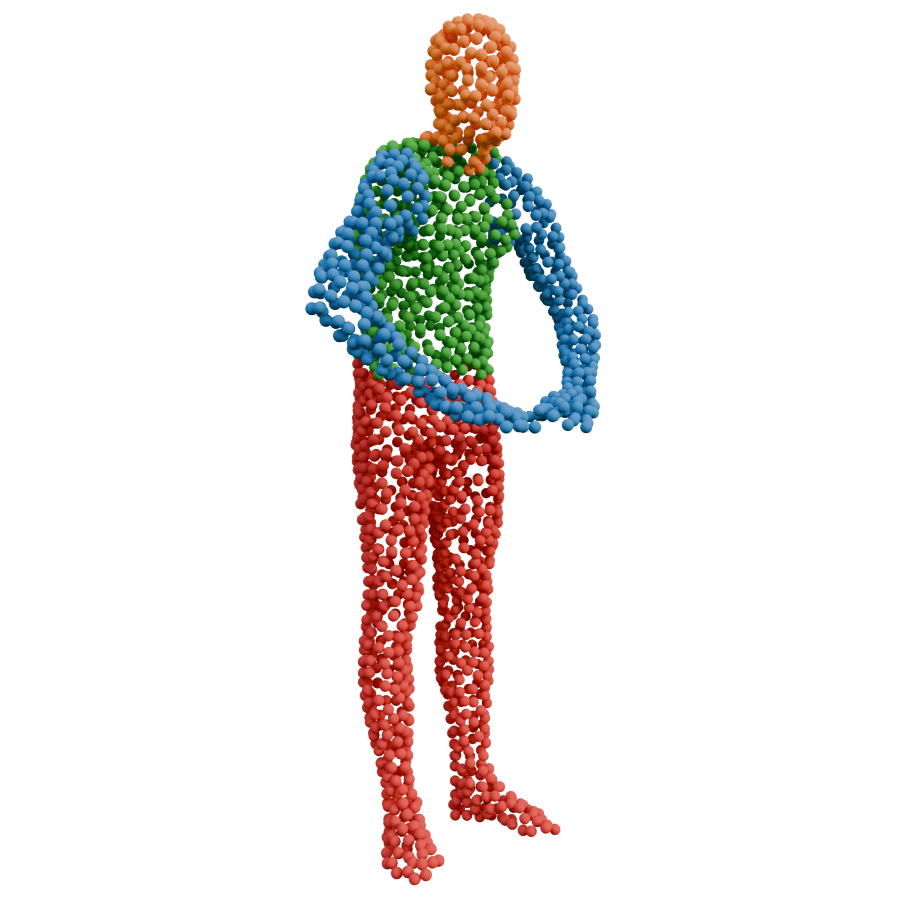} &
    \includegraphics[width=0.30\linewidth]{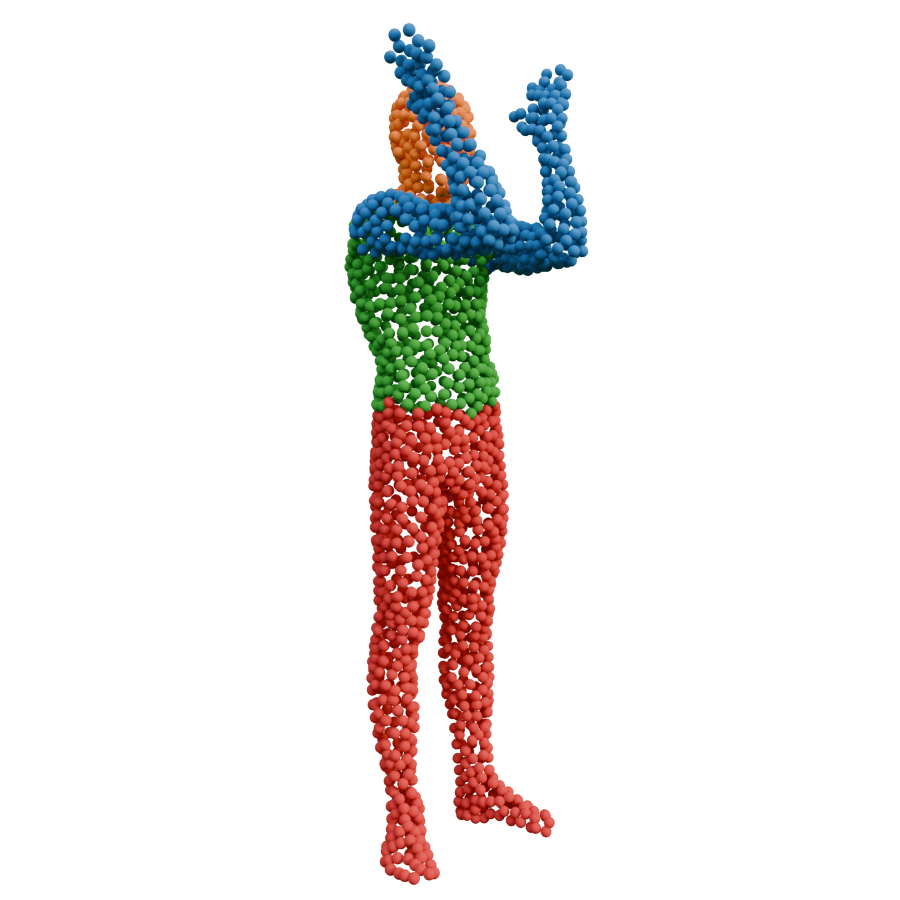} &
    \includegraphics[width=0.30\linewidth]{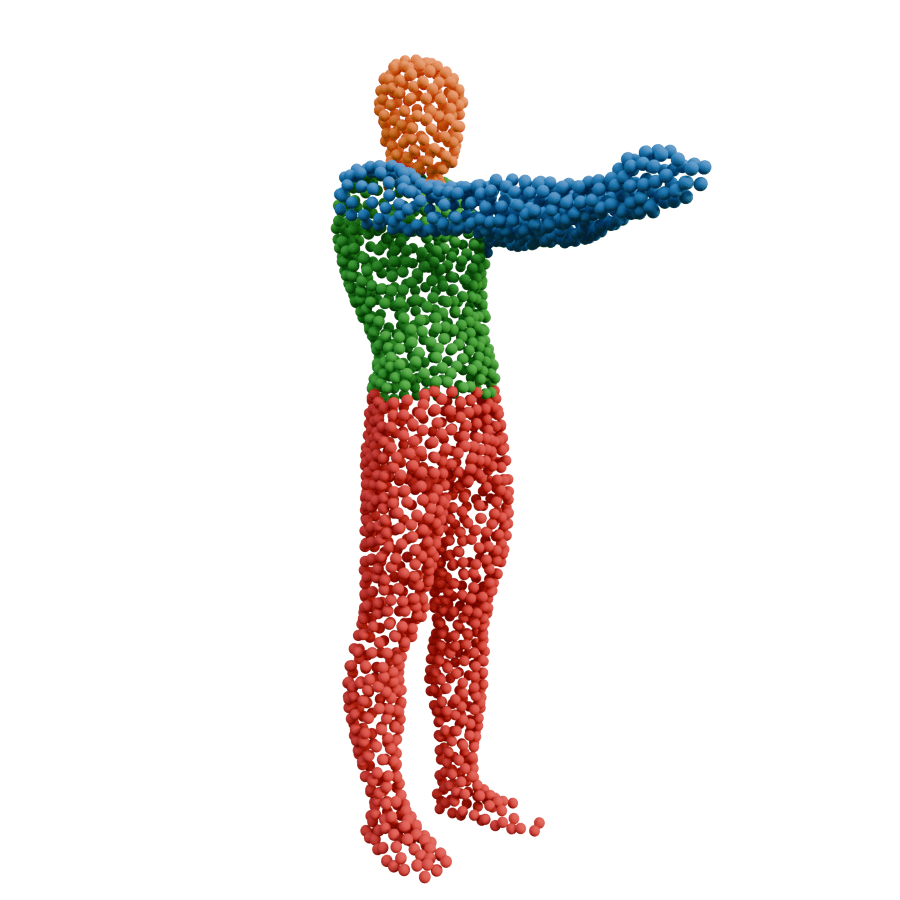} \\[4pt]

    \rowlabelnew{COPS} &
    \includegraphics[width=0.30\linewidth]{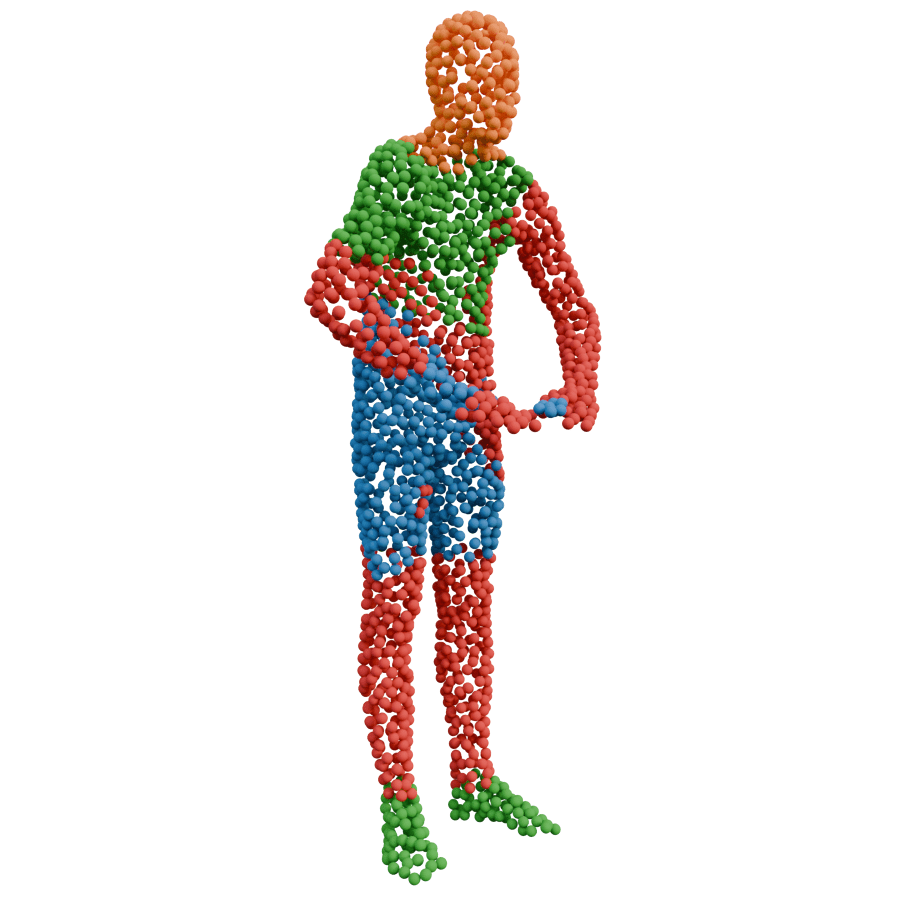} &
    \includegraphics[width=0.30\linewidth]{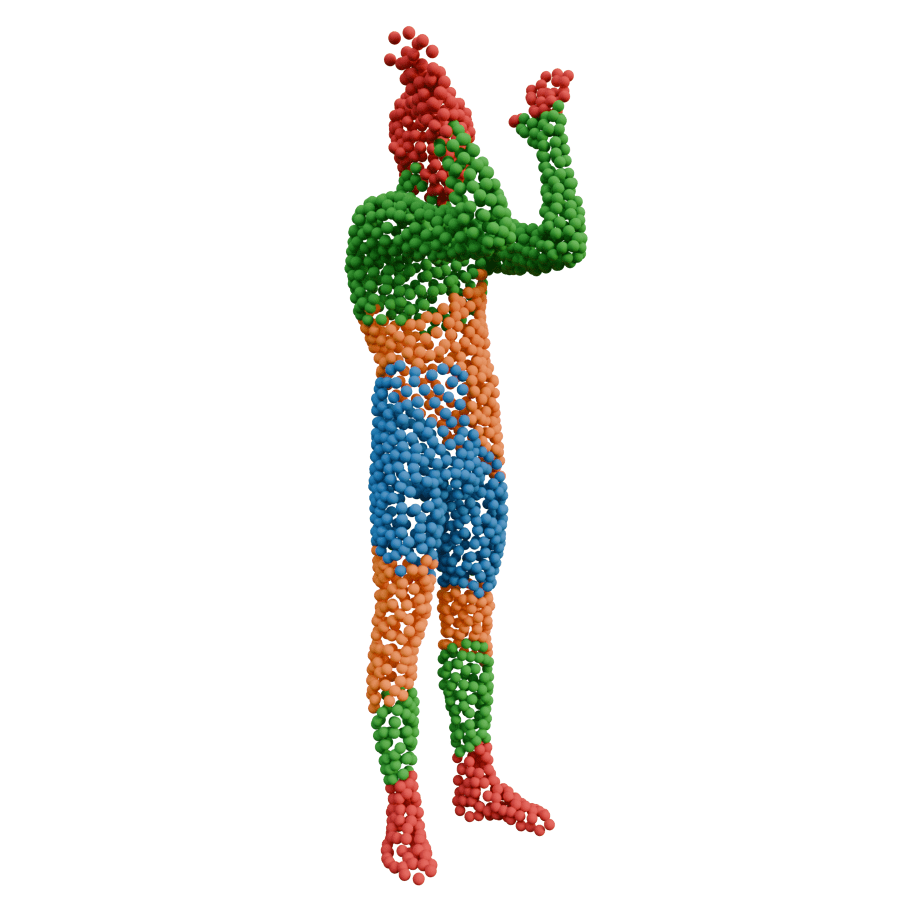} &
    \includegraphics[width=0.30\linewidth]{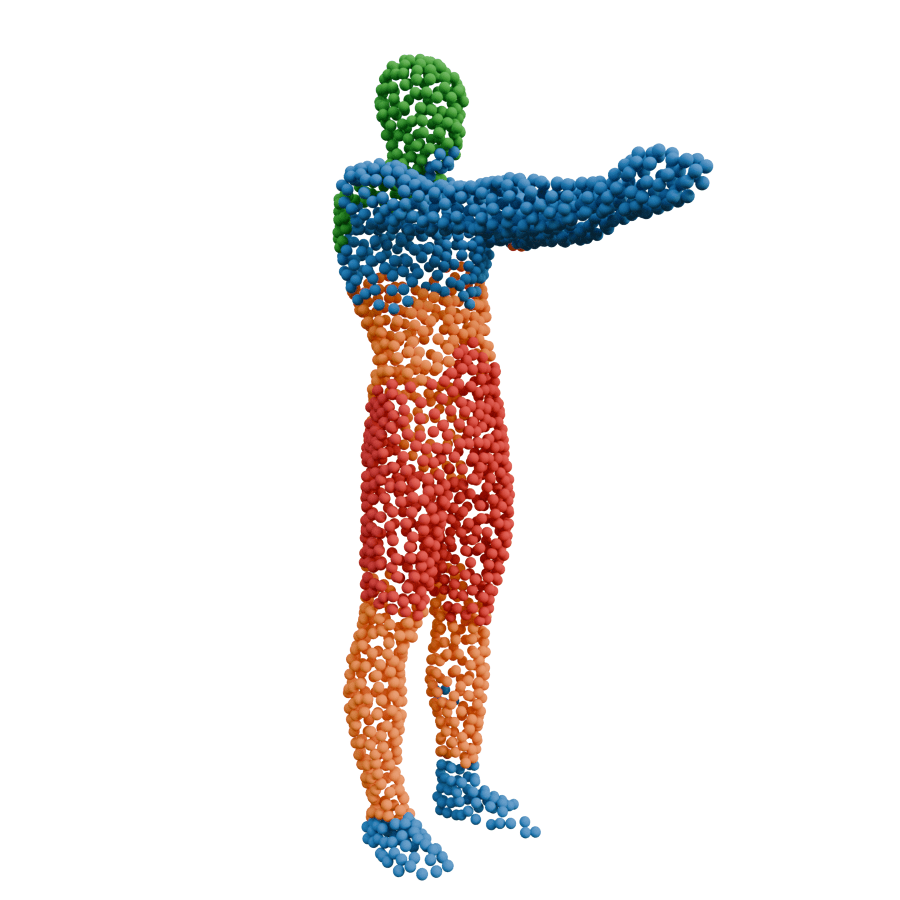} \\[4pt]

    \rowlabelnew{Find3D} &
    \includegraphics[width=0.30\linewidth]{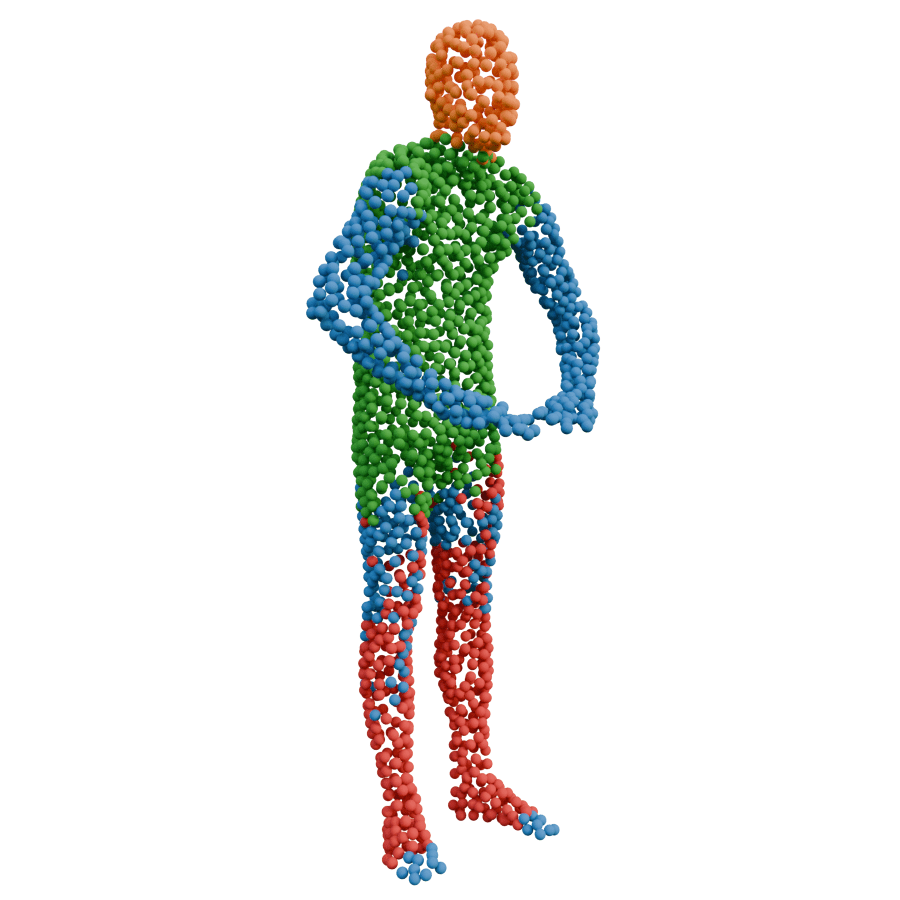} &
    \includegraphics[width=0.30\linewidth]{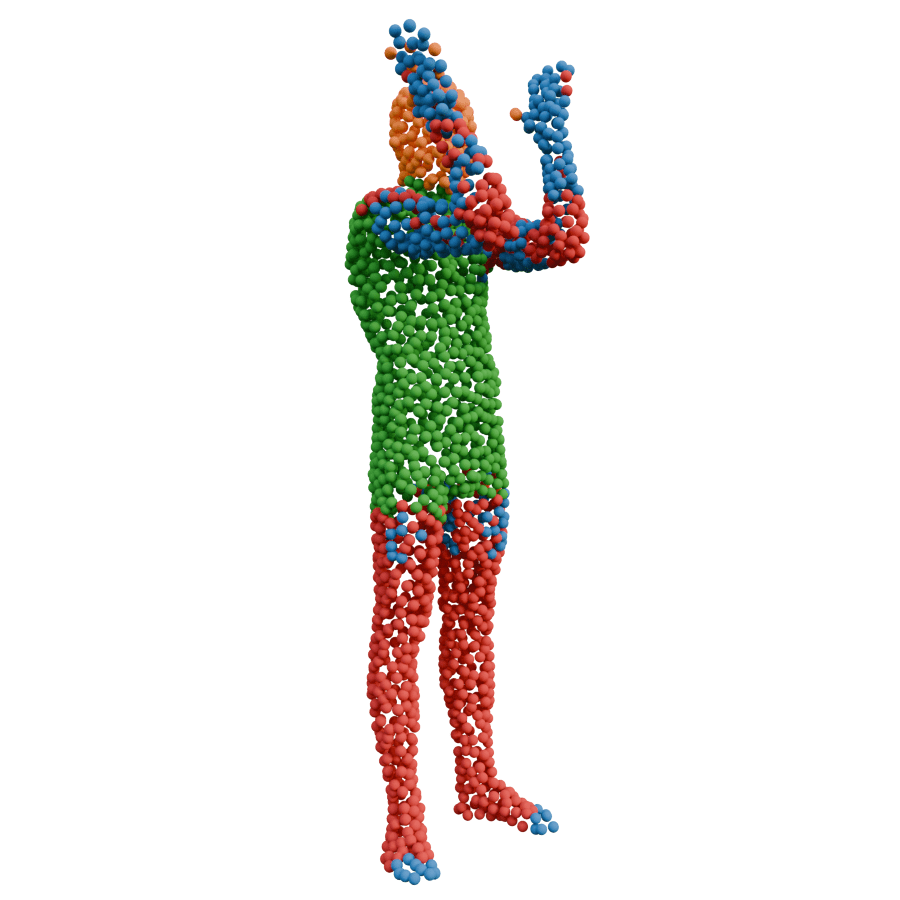} &
    \includegraphics[width=0.30\linewidth]{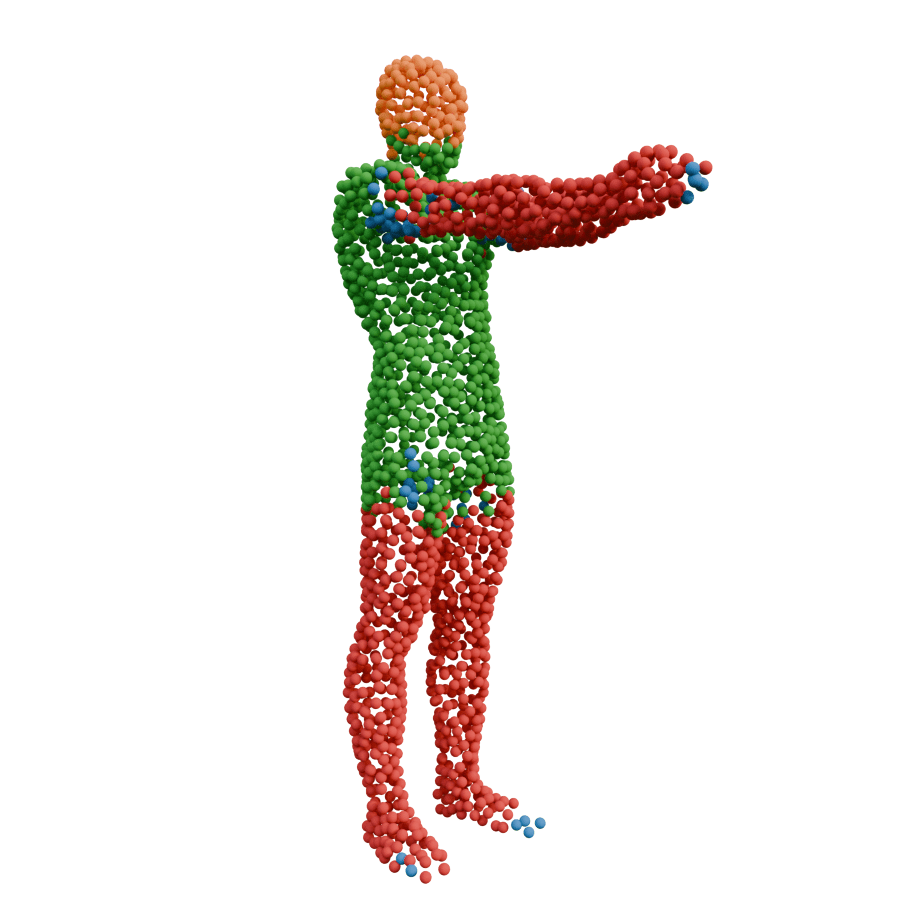} \\[4pt]

    \rowlabelnew{PatchAlign3D} &
    \includegraphics[width=0.30\linewidth]{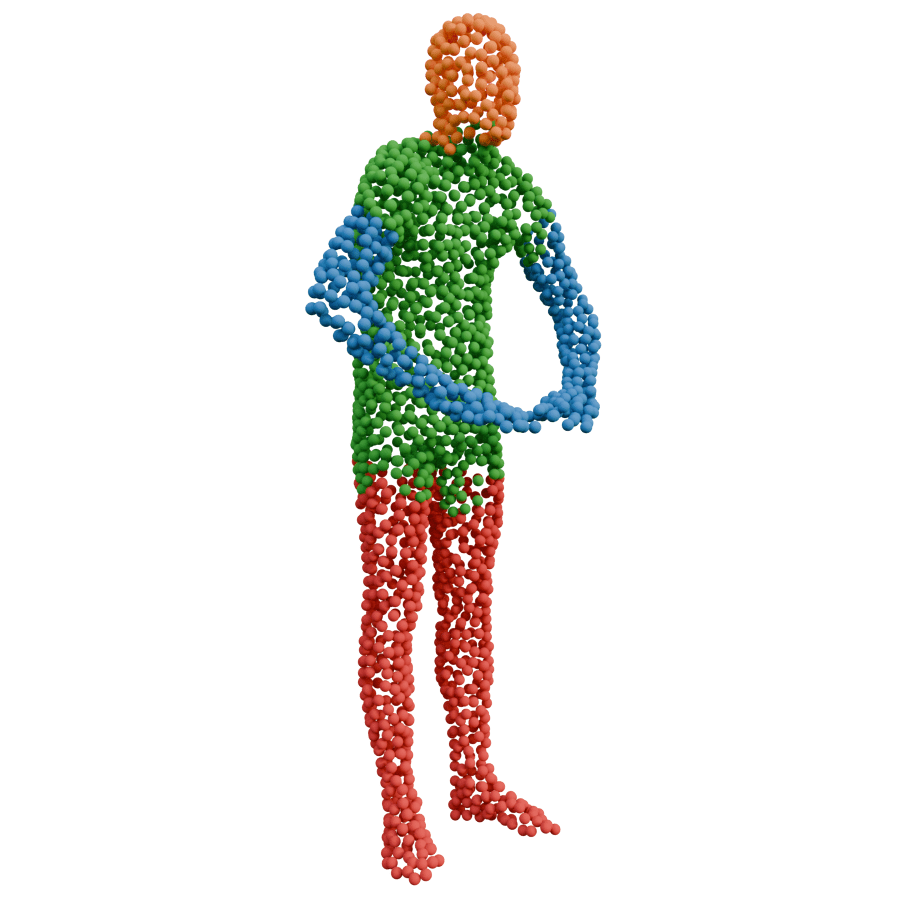} &
    \includegraphics[width=0.30\linewidth]{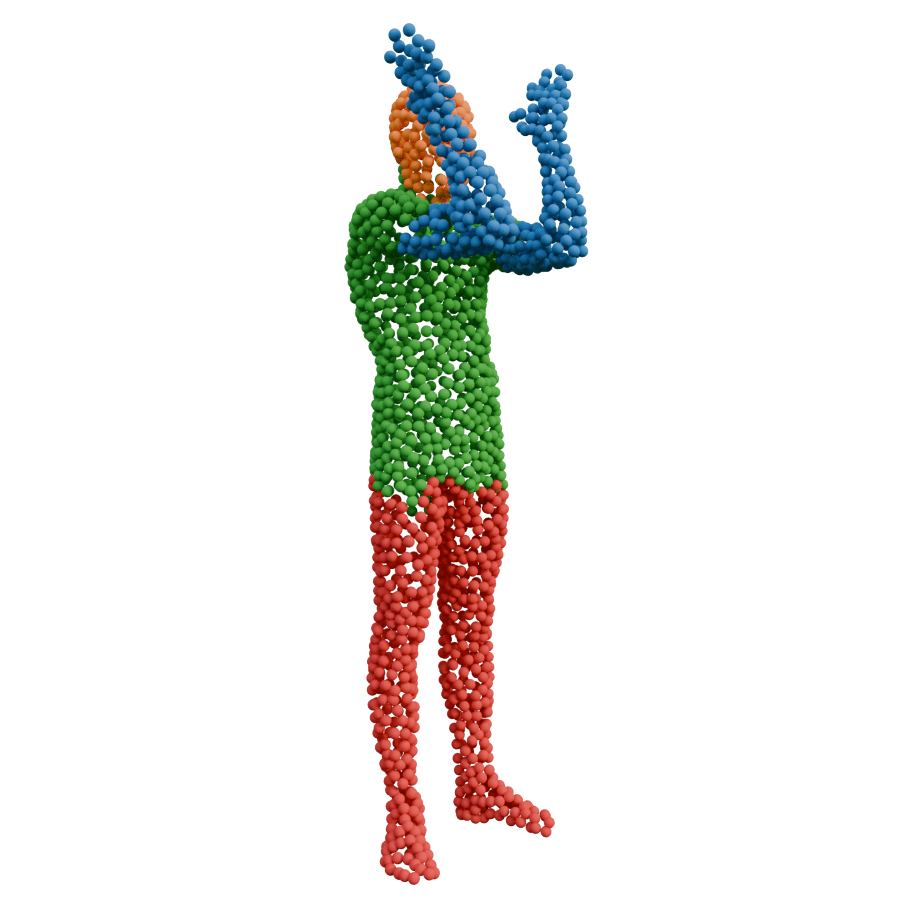} &
    \includegraphics[width=0.30\linewidth]{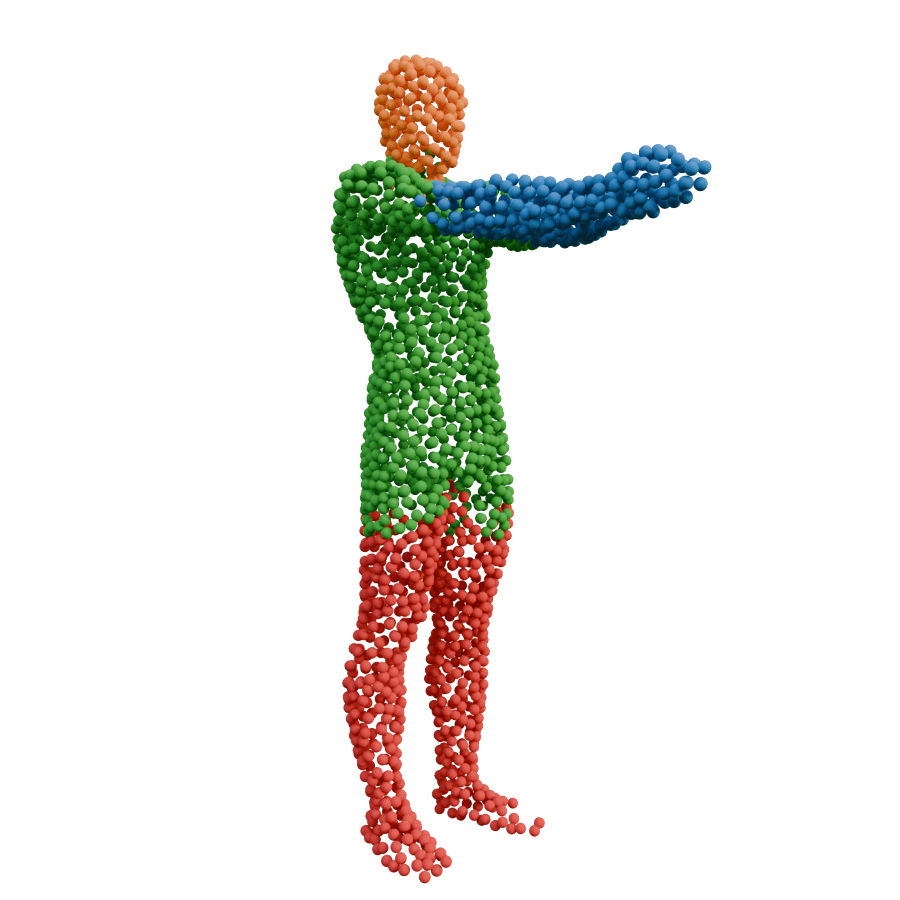} \\[2pt]

    & \multicolumn{3}{c}{\scriptsize
      \{\textcolor{blue}{arm}, \textcolor{orange}{head}, \textcolor{red}{leg}, \textcolor{darkgreen}{torso}\}} \\
\end{tabular}

\caption{\textbf{Qualitative comparison on non-rigid human shapes from FAUST \cite{bogo2014faust, abdelreheem2023satr}.}
We show ground truth and predictions from COPS \cite{garosi20253d}, Find3D \cite{ma2025find}, and PatchAlign3D across three
representative shapes (columns). The part legend below specifies the semantic labels
used for zero-shot prediction. PatchAlign3D produces cleaner segmentations than prior
methods and is less noisy than Find3D's encoder–decoder outputs.}

    \label{fig:humans_3x4}
\end{figure}

\mypara{Metrics.}
We report two aggregated segmentation metrics: mean Intersection-over-Union over instances (mIoU) and mean category-wise Intersection-over-Union (cIoU). 
In our setting, mIoU averages part-wise IoU over all test shapes, giving equal weight to each instance regardless of category frequency. 
cIoU averages IoU across object categories before aggregation, making it less sensitive to dataset imbalance. 
Both metrics capture the quality and consistency of part predictions in a zero-shot context.

\mypara{Baselines.}
We compare PatchAlign3D against both rendering-based vision–language methods and feed-forward 3D encoders.
Among rendering-based approaches, we include PointCLIPv2 \cite{zhu2022pointclip}, which leverages CLIP \cite{radford2021learning} features, and COPS \cite{garosi20253d}, which uses dense DINOv2 \cite{oquab2023dinov2} features. 
As a feed-forward baseline, we evaluate Find3D~\cite{ma2025find}, an encoder–decoder architecture trained on the same dataset as our approach and designed to produce point-level features aligned with language.
For reference, we also report the mesh-based methods SATR \cite{abdelreheem2023satr} and 3DH \cite{decatur20233d}, although they use high-resolution meshes rather than point clouds and are therefore not strictly comparable. 
Additionally, we include PartSLIP \cite{liu2023partslip} on the PartNetE benchmark, where it performs strongly, but its inference cost (several minutes per shape) makes evaluation on all benchmarks impractical. 
When baseline results were unavailable, we re-ran models using their official code. 
Finally, to avoid prompt engineering bias, we follow a unified template ("\{part\}", "a \{part\}", "\{part\} part") for all textual descriptions and re-run experiments when necessary to ensure consistency.

\mypara{Implementation settings.}
For Stage 1 distillation, we extract visual features using DINOv2 \cite{oquab2023dinov2} to remain consistent
with prior work such as COPS. We ablate alternative visual encoders in the supplementary material. For Stage 2 and inference, we employ the widely used OpenCLIP ViT-bigG-14 text encoder~\cite{radford2021learning}, following recent 3D foundation models~\cite{liu2023openshape, zhou2023uni3d}. An ablation of different text encoders is provided in the supplementary material.

Throughout both training and inference, PatchAlign3D uses only XYZ coordinates as input to point clouds of size 2048. The number of patches is fixed to 128, and each patch contains 32 points.
In contrast, several baselines (e.g., Find3D) incorporate richer input modalities such as
RGB colors and surface normals whenever available.

We train both Stage 1 and Stage 2 for 100 epochs with a batch size of 32 using the
same training set. In Stage 2, we fine-tune only the last transformer block and the
projector, leaving the remaining layers frozen. Additional ablations on patch granularity, prompt robustness, and training design are provided in the supplementary material. We implement the whole framework in PyTorch \cite{paszke2019pytorch}.


\begin{table}[t]
\centering
\resizebox{\columnwidth}{!}{%
\begin{tabular}{l|l|c|cccc}
\hline
\textcolor{gray}{Pipeline} & Method & mIoU & Arm & Head & Leg & Torso \\ \hline
\multicolumn{7}{c}{Mesh methods} \\ \hline
\textcolor{gray}{Rendering} & 3DH~\cite{decatur20233d}  & 16.5 & 28.6 & 14.2 & 14.9 & 8.2 \\
\textcolor{gray}{Rendering} & SATR~\cite{abdelreheem2023satr}\,*  & 59.6 & 44.8 & 73.2 & 65 & 55.4 \\
\hline
\multicolumn{7}{c}{Point-cloud methods} \\ \hline
\textcolor{gray}{Rendering} & PointCLIPv2~\cite{zhu2022pointclip}  & 13.6 & 14.0 & 15.8 & 24.1 & 0.3 \\
\textcolor{gray}{Rendering} & COPS~\cite{garosi20253d} & 30.4 & 29.8 & 33.4 & 48.2 & 10.2 \\
\textcolor{gray}{Feed-forward} & Find3D~\cite{ma2025find} & 63.2 & 59.8 & 81.0 & 59.9 & \textbf{52.0} \\
\rowcolor{OursTint}
\textcolor{gray}{Feed-forward} & PatchAlign3D & \textbf{67.8} & \textbf{68.7} & \textbf{86.4} & \textbf{65.5} & 49.7 \\ \hline
\end{tabular}%
}
\caption{\textbf{Zero-shot part segmentation results on FAUST \cite{bogo2014faust, abdelreheem2023satr}.} Our main comparison is with point-cloud methods, although we also report mesh-based approaches for completeness. *For fairness, 
the SATR input is a mesh reconstructed from a 5000-point point cloud. 
PatchAlign3D achieves the best performance and significantly surpasses rendering-based baselines.}
\label{tab:faust}
\end{table}

\subsection{Quantitative Results}

\cref{tab:shapenetpart} reports results on ShapeNetPart, where PatchAlign3D
establishes a new state of the art in open-world segmentation. Our method surpasses the strongest prior approach,
COPS, by \textbf{+31.3\%} mIoU and \textbf{+20.9\%} cIoU, and achieves
consistent gains across 15 of the 16 object categories. It also substantially
outperforms its feed-forward counterpart, Find3D, despite both methods being trained
on the same data. These results indicate that PatchAlign3D's training pipeline produces significantly stronger local representations, even though it operates on coarser patch features.
\cref{tab:faust} shows results on FAUST's non-rigid human shapes.
PatchAlign3D again outperforms existing point-cloud methods, improving over Find3D by
\textbf{+4.6\%} mIoU.
On PartNetE (\cref{tab:partnete_scanobjectnn}), we assign a
"body" label to unlabeled points since our approach relies on patch-text similarity. Despite these constraints and the dataset's
fine-grained part definitions, PatchAlign3D maintains a clear margin over previous baselines.
ScanObjectNN introduces substantial domain shift due to real-world noise and background
clutter. As shown in \cref{tab:partnete_scanobjectnn}, PatchAlign3D still achieves
the best performance, with improvements of \textbf{+3.9\%} mIoU and \textbf{+4.3\%} cIoU over
the strongest competitor.
On the Objaverse–General benchmark (\cref{tab:objaverse_general}), our
method exceeds Find3D on both seen and unseen categories, demonstrating strong
generalization under category shift. Overall, PatchAlign3D consistently outperforms 
rendering-based methods and even the best feed-forward baseline, despite being trained on
exactly the same data.

\mypara{Inference speed.}
\cref{tab:inference_speed} compares the runtime of PatchAlign3D with rendering-based and feed-forward baselines.
PatchAlign3D achieves inference speed on par with existing feed-forward models and is faster than rendering-based approaches such as COPS. This efficiency comes from our single-pass architecture and is a key step toward real-time 3D part segmentation.

\begin{table}[t]
\centering
\tiny
\resizebox{\columnwidth}{!}{%
\begin{tabular}{lll|ll}
\hline
\multirow{2}{*}{Method} 
& \multicolumn{2}{c}{PartNetE}
& \multicolumn{2}{c}{ScanObjectNN} \\[-2pt]

\cmidrule(lr){2-3}
\cmidrule(lr){4-5}

& mIoU & cIoU & mIoU & cIoU \\ \hline

PointCLIPv2~\cite{zhu2022pointclip} & 23.8    & 26.0  &   9.0   &   11.0   \\
PartSLIP~\cite{liu2023partslip}    & --    & 36.4 & --   & --   \\
COPS~\cite{garosi20253d}        & 27.0 & 29.3 &   17.7   &   20.2   \\
Find3D~\cite{ma2025find}      & 16.4 & 17.1    &  18.8    &   21.0  \\

\rowcolor{OursTint}
PatchAlign3D &   \textbf{41.4}   &   \textbf{42.2}    &  \textbf{22.7}    &    \textbf{25.3}  \\ \hline
\end{tabular}%
}
\caption{\textbf{Zero-shot shape part segmentation on PartNetE \cite{liu2023partslip} and ScanObjectNN \cite{uy2019revisiting}.} 
PointCLIPv2, COPS, and Find3D are evaluated using part labels only. 
We also include the reported PartSLIP result on PartNetE, where it performs best, 
but do not re-evaluate it on ScanObjectNN due to its high computational cost 
(about 4 minutes per shape). PatchAlign3D achieves the highest performance across both datasets.}
\label{tab:partnete_scanobjectnn}
\end{table}

\begin{table}[t]
\centering
\small
\begin{tabular}{l|c|c}
\hline
\multirow{2}{*}{Method} 
& \multicolumn{2}{c}{Objaverse–General} \\
\cmidrule(lr){2-3}
& \shortstack[c]{Seen\\\textcolor{gray}{\scriptsize mIoU}}
& \shortstack[c]{Unseen\\\textcolor{gray}{\scriptsize mIoU}} \\ \hline
Find3D~\cite{ma2025find} & 28.9 & 34.6 \\
\rowcolor{OursTint}
PatchAlign3D & \textbf{37.49} & \textbf{35.61} \\ \hline
\end{tabular}
\caption{\textbf{Zero-shot part segmentation on Objaverse–General.}
Because the official split is not provided, we define a seen/unseen split with 
14 unseen categories. PatchAlign3D achieves the best results on both splits and 
shows strong robustness to category shift.}
\label{tab:objaverse_general}
\end{table}

\subsection{Qualitative Results}

\cref{fig:qual_4x6} presents qualitative comparisons on six representative
ShapeNetPart categories. We include the strongest rendering-based (COPS) and 
feed-forward (Find3D) baselines. PatchAlign3D consistently produces sharper and more 
faithful segmentations across a wide range of geometries. Although our method operates 
on patches, the predicted part boundaries remain well aligned with the underlying 
surfaces, and the resulting segments are spatially coherent rather than fragmented. 
In contrast, Find3D often exhibits point-level noise, with neighbouring points switching 
labels abruptly.

\cref{fig:humans_3x4} shows results on FAUST's non-rigid human shapes using coarse part 
labels. Despite operating on sparse point samples, PatchAlign3D maintains high 
segmentation quality and produces markedly cleaner predictions than prior methods. 
The rendering-based COPS baseline degrades significantly under deformation, whereas 
PatchAlign3D remains stable and is less affected by pose variation.

\begin{figure}[t]
    \centering
    \setlength{\tabcolsep}{1pt}
    \renewcommand{\arraystretch}{0}

    \begin{tabular}{@{}ccc@{}}
        \includegraphics[width=0.31\linewidth]{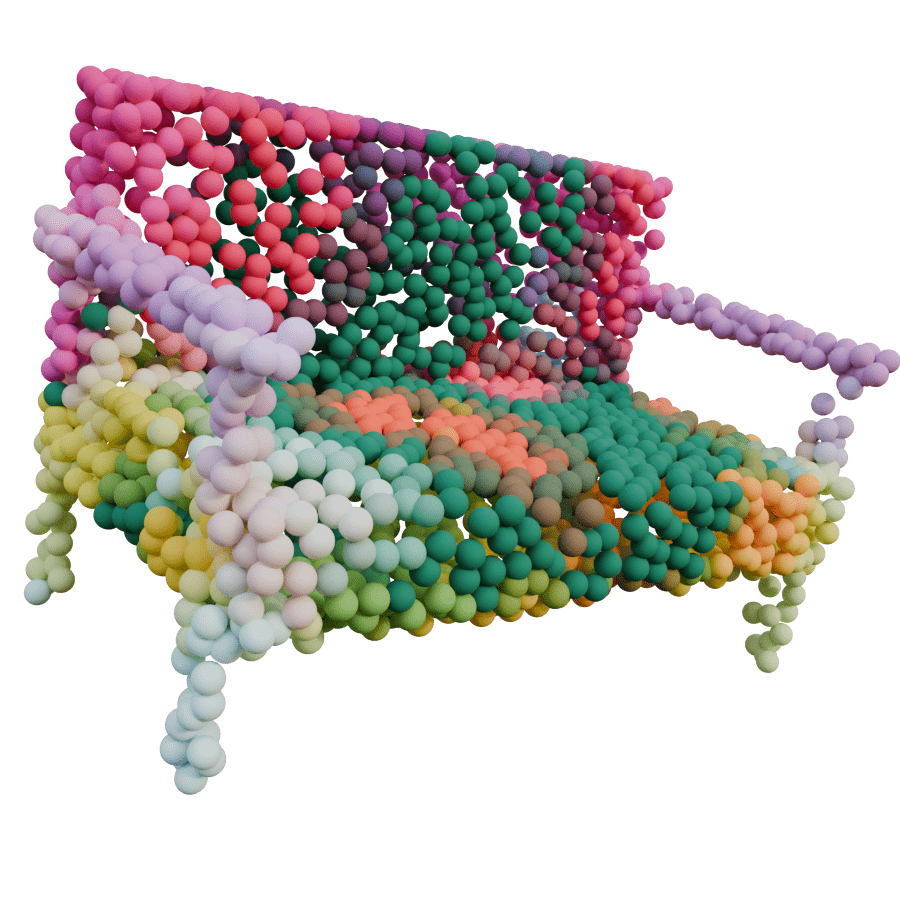} &
        \includegraphics[width=0.31\linewidth]{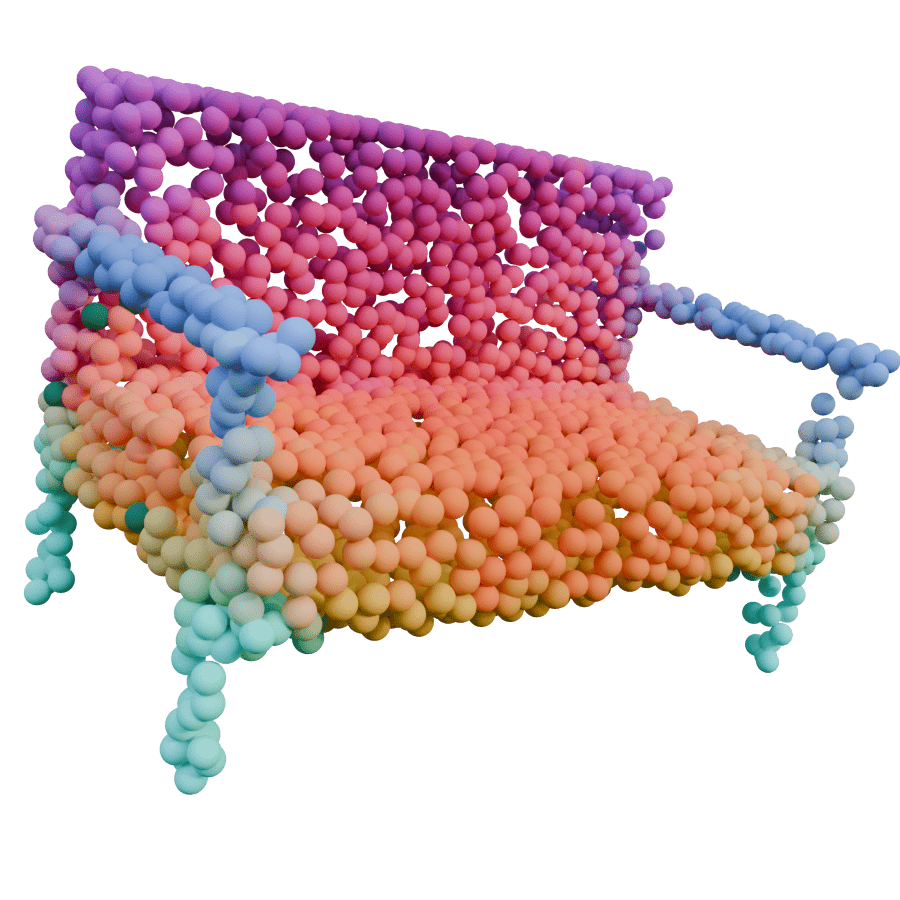} &
        \includegraphics[width=0.31\linewidth]{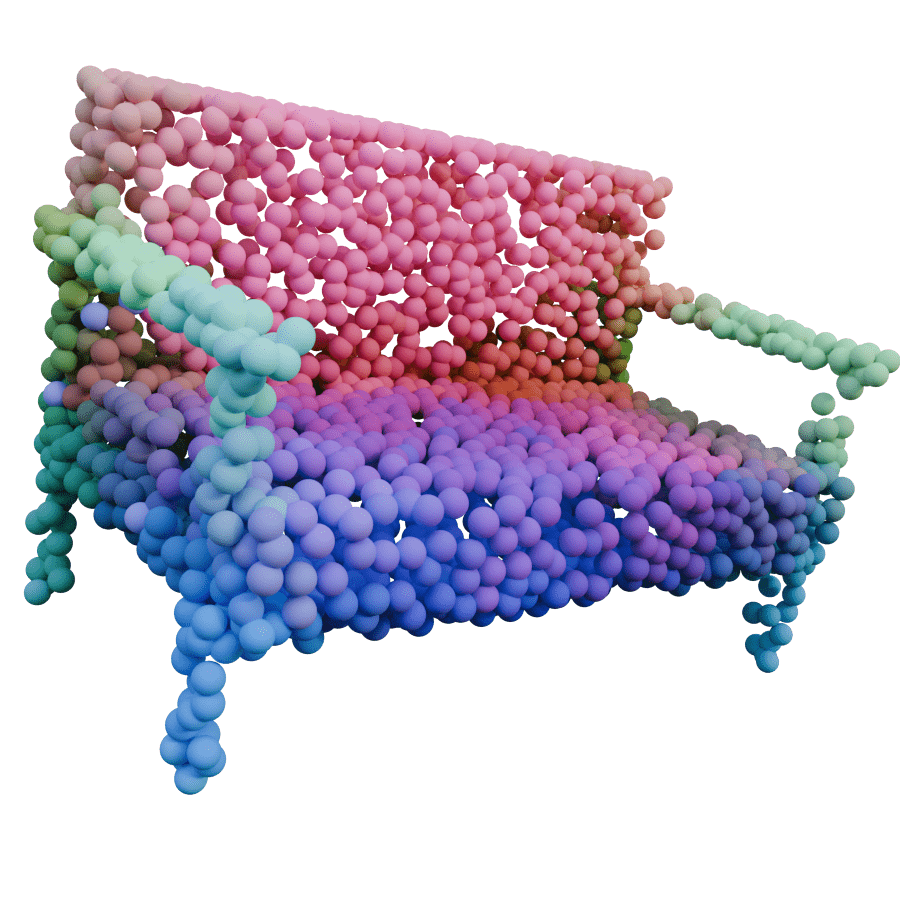} \\[-2pt]

        \includegraphics[width=0.31\linewidth]{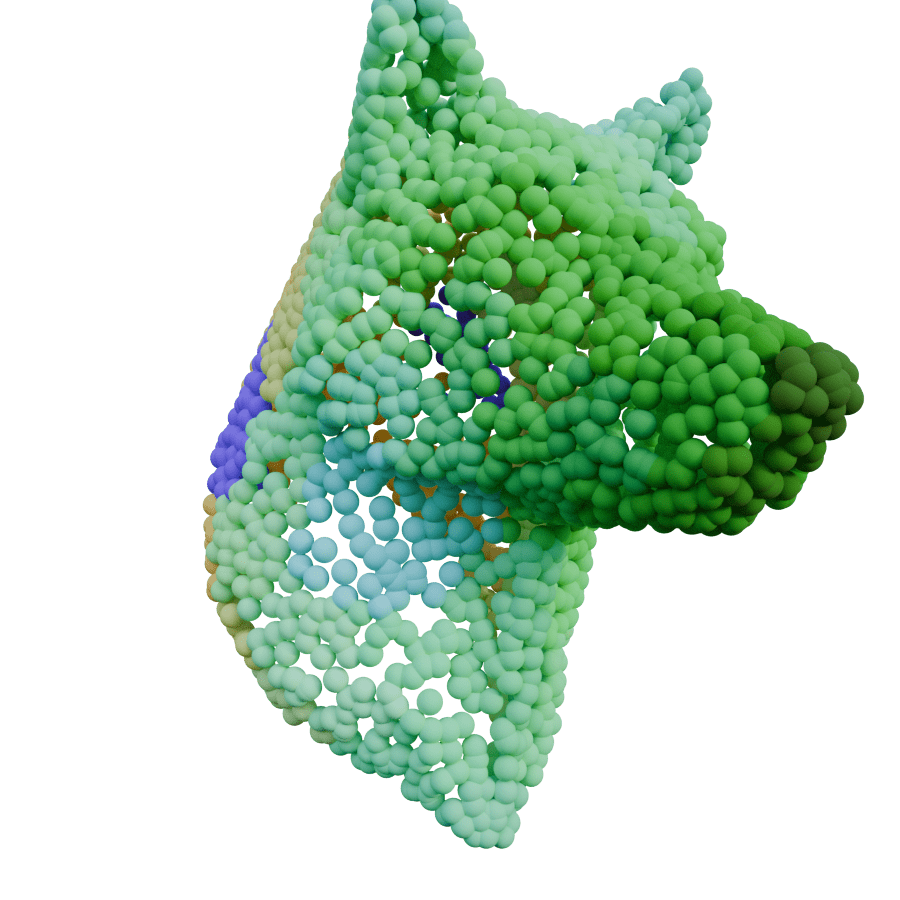} &
        \includegraphics[width=0.31\linewidth]{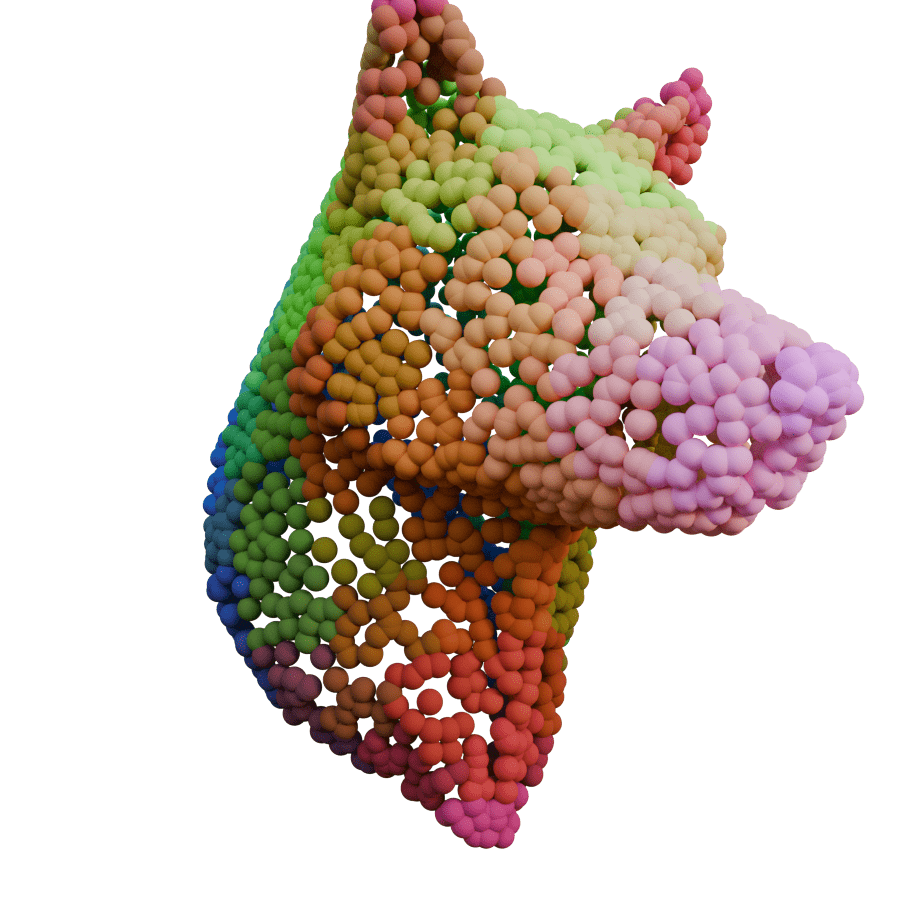} &
        \includegraphics[width=0.31\linewidth]{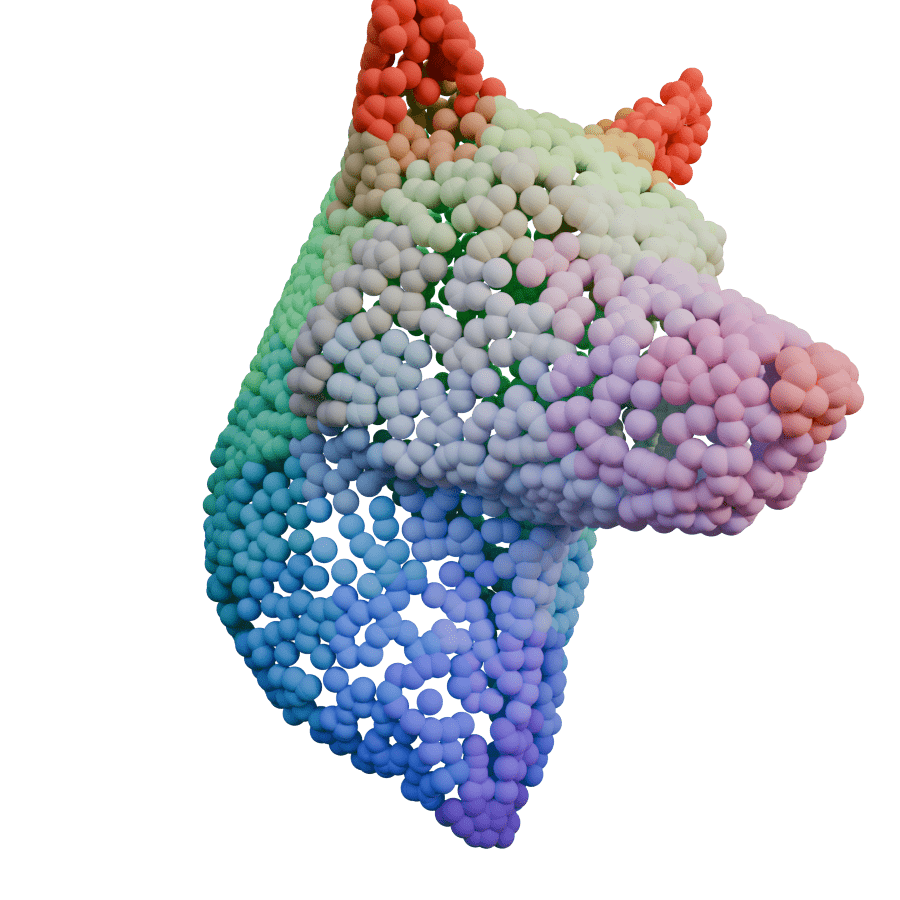} \\[3pt]

        {\scriptsize (a) DINOv2} &
        {\scriptsize (b) Stage 1} &
        {\scriptsize (c) Stage 1 + Stage 2} \\
    \end{tabular}

    \caption{\textbf{Feature comparison across stages.}
    We visualize features from DINOv2, Stage 1, and Stage 1 + Stage 2 of our approach
    on example point clouds from the validation split of the training data.
    Stage 1 refines DINOv2 features, and Stage 2 further preserves them while assigning downstream text capabilities.}
    \label{fig:stage_features}
\end{figure}

\subsection{Ablation Study}

\mypara{Multi-stage pre-training strategy.}
\cref{tab:ablation_stage} examines the contribution of our two-stage pre-training framework. We observe that \textbf{Stage 2 alone}, which performs multi-positive contrastive alignment between patch tokens and text embeddings, already achieves state-of-the-art zero-shot performance on the ShapeNetPart segmentation benchmark. This indicates that our patch-wise contrastive formulation is sufficiently effective to learn semantically meaningful local features even without dense 2D supervision.
Nevertheless, initializing Stage 2 from the geometry-aware representations learned during \textbf{Stage 1} yields a clear improvement.

We also validate our two-stage approach by evaluating a \textbf{joint-training} variant that jointly optimizes both losses using each stage's respective projection head. This approach performs slightly worse than Stage 2 alone. We attribute this degradation to the fundamentally different nature of the two objectives: when applied concurrently, these competing signals interfere with each other. The clear gains of our two-stage design therefore, show the importance of decoupling dense feature distillation from text alignment.

\mypara{Qualitative comparison of learned features.}
\cref{fig:stage_features} visualizes the representations produced by the 2D DINOv2 backbone, our Stage 1 encoder, and the final PatchAlign3D model (Stage 1 + Stage 2). For comparison, we project each embedding space independently to RGB using PCA. DINOv2 features are lifted to the point cloud via the same multi-view back-projection used during preprocessing, whereas Stage 1 and Stage 2 features originate from patch tokens and are assigned to points through nearest-centroid propagation. While DINOv2 offers a strong initialization, its back-projected features remain noisy and exhibit inconsistencies across the surface. Stage 1 yields notably more coherent, geometry-aware patterns: different semantic regions (e.g., the seat vs. the back of a sofa, or the nose, ears, and neck of a wolf head) form clearly separated clusters. Stage 2 preserves this structural organization and further aligns language, endowing the features with open-vocabulary semantic behavior. Together, these results show that Stage 1 successfully refines dense 2D priors into stable 3D patch embeddings, and that Stage 2 enhances them with the text-driven semantics required for zero-shot part labeling. Additional qualitative comparisons are provided in the supplementary material.

\begin{table}[t]
\centering
\small
\resizebox{\columnwidth}{!}{%
\begin{tabular}{l|c|c|c}
\hline
Method & Modality & Type & Inference Time (s) \\ \hline
SATR          & Mesh & Rendering & 111 \\
COPS          & Point Cloud  & Rendering & 1.38 \\
PointCLIPv2   & Point Cloud  & Rendering & 1.20 \\
Find3D        & Point Cloud  &  Feed-forward  & \textbf{0.4} \\
\rowcolor{OursTint}
PatchAlign3D          & Point Cloud  &  Feed-forward  & \textbf{0.4} \\ \hline
\end{tabular}%
}
\caption{\textbf{Inference speed comparison.} PatchAlign3D matches the efficiency of the feed-forward approaches and is faster than rendering-based methods.}
\label{tab:inference_speed}
\end{table}

\begin{table}[t]
\centering
\small
\resizebox{0.6\columnwidth}{!}{%
\begin{tabular}{l|cc}
\hline
Configuration & mIoU & cIoU \\ \hline
Stage 2 only & 50.5 & 50.0 \\
Joint training & 50.2 & 48.6 \\
\rowcolor{OursTint}
2-Stage (PatchAlign3D) & \textbf{56.9} & \textbf{53.1} \\ \hline
\end{tabular}%
}
\caption{\textbf{Ablation on multi-stage training strategy.}
While Stage 2 alone already surpasses prior baselines, our full two-stage training
further improves performance. Jointly optimizing both stages at once
degrades pre-training. Evaluation is done on ShapeNetPart.}
\label{tab:ablation_stage}
\end{table}

\section{Conclusion}
\label{sec:conclusion}

We introduced PatchAlign3D, an encoder-only 3D transformer that learns semantically meaningful patch-level representations for zero-shot part segmentation. Our two-stage pre-training strategy first distills dense 2D priors from a vision backbone into 3D patch tokens, and then aligns these tokens with a text encoder using a multi-positive contrastive objective. This design enables feed-forward, single-pass inference at test time, without any multi-view rendering or prompt engineering. Across five benchmarks spanning synthetic and real-world data, rigid and non-rigid shapes, and seen and unseen categories, PatchAlign3D consistently outperforms both rendering-based vision--language pipelines and previous feed-forward 3D encoders, while remaining computationally efficient.

\mypara{Limitations and future work.}
Despite these gains, our approach has several limitations. PatchAlign3D is pre-trained on a curated Objaverse subset with imperfect pseudo-part annotations derived from SAM and a language model. The pre-training set still covers only a small fraction of the 800K+ objects available in Objaverse.
These constraints open up several promising directions for future work. One avenue is to scale PatchAlign3D to larger, more diverse data sets, while increasing the encoder's scale. Another direction is to replace the fixed patching scheme with more adaptive partitioning strategies to handle point clouds of different sizes. A natural extension is to generalize the encoder to inherit the global understanding of existing 3D foundation models.

\mypara{Acknowledgements} Parts of this work were supported by the ERC Consolidator Grant 101087347 (VEGA). The authors also gratefully acknowledge gifts from Ansys and Adobe Inc. This work was also supported by funding from King Abdullah University of Science and Technology (KAUST) – Center of Excellence for Generative AI, under award number 5940 and a gift from Google.

{\small
\bibliographystyle{ieeenat_fullname}
\bibliography{11_references}
}

\ifarxiv \clearpage \appendix 
\setcounter{figure}{5}
\setcounter{table}{6} 

This supplementary document complements the main manuscript by providing expanded quantitative analyses and qualitative visualizations of PatchAlign3D. First, Section A presents comprehensive ablation studies that validate our architectural design choices, specifically evaluating the impact of different dense 2D visual encoders, text encoders, Stage 2 freezing strategies, patch granularity and its impact on boundary performance, prompt sensitivity and negative sampling strategy. We also provide additional qualitative comparisons in Section B, illustrating the robustness of our patch-level alignment in both text-to-feature and anchor-based scenarios. Finally, we detail our experimental framework for zero-shot and few-shot keypoint detection in Section C, demonstrating the method's fine-grained capabilities and potential for additional applications.

\section{Additional Ablations}

\paragraph{Ablation on the 2D visual encoder.}
\cref{tab:2d_encoder_ablation} shows that PatchAlign3D is compatible with any
visual encoder that produces dense spatial features. ViT-based~\cite{dosovitskiy2020image} models such as
DINOv1~\cite{caron2021emerging}, CLIP~\cite{radford2021learning}, and DINOv2~\cite{oquab2023dinov2} all yield strong results, demonstrating that Stage 1
distillation does not depend on a specific visual backbone. Dense representation
learning appears particularly important: DINOv1~\cite{caron2021emerging} and DINOv2~\cite{oquab2023dinov2}, both trained with
dense objectives, outperform CLIP. Surprisingly, DINOv3~\cite{simeoni2025dinov3} severely underperforms, possibly
because its features are optimized for more fine-grained objectives and are
less suitable for shape segmentation, showing that the passage from DINOv2 to DINOv3 is not necessarily beneficial for all downstream tasks. These results justify our
use of DINOv2~\cite{oquab2023dinov2} for the main experiments.

\paragraph{Ablation on the text encoder.}
As shown in \cref{tab:text_encoder_ablation}, PatchAlign3D performs well with
a range of text encoders. The CLIP ViT-bigG~\cite{radford2021learning} model achieves the best results,
likely due to its large-scale multimodal training. However, the performance of
Gemma-2-9B-it~\cite{team2024gemma}, despite being trained purely on text, is notable and indicates that Stage 2 learning does not rely on a
visually-grounded text tower. This highlights the robustness of our
multi-positive patch-level alignment mechanism.

\begin{table}[t]
    \centering
    \begin{tabular}{lcc}
        \toprule
        2D encoder & mIoU & cIoU \\
        \midrule
        DINOv1~\cite{caron2021emerging} & 51.82 & 54.39 \\
        DINOv3~\cite{simeoni2025dinov3} & 46.52 & 42.76 \\
        OpenCLIP ViT-bigG-14 ~\cite{radford2021learning}& 49.32 & 52.96 \\
        DINOv2~\cite{oquab2023dinov2}(ours) & \textbf{56.90} & \textbf{53.10} \\
        \bottomrule
    \end{tabular}
    \caption{\textbf{Ablation on the 2D encoder used during Stage 1.}
    We evaluate several dense visual encoders for multi-view 2D feature distillation.
    All models produce competitive results, but dense-trained ViTs~\cite{dosovitskiy2020image} such as
    DINOv1 and DINOv2 perform best, supporting our choice of DINOv2. Evaluation is done on ShapeNetPart.}
    \label{tab:2d_encoder_ablation}
\end{table}

\begin{table}[t]
    \centering
    \begin{tabular}{lcc}
        \toprule
        Text encoder & mIoU & cIoU \\
        \midrule
        SigLIP~\cite{zhai2023sigmoid} & 46.44 & 40.51 \\
        OpenCLIP ViT-bigG-14~\cite{radford2021learning} (ours) & \textbf{56.90} & \textbf{53.10} \\
        Gemma-2-9B-it~\cite{team2024gemma}& 54.83 & 50.98 \\
        \bottomrule
    \end{tabular}
    \caption{\textbf{Ablation on the text encoder.}
    All text encoders yield substantial improvements over prior baselines.
    CLIP ViT-bigG~\cite{radford2021learning} remains the strongest, but even purely textual encoders such as
    Gemma-2-9B-it~\cite{team2024gemma} perform surprisingly well, indicating that Stage 2 does not strictly
    require a vision–language pre-trained text tower. Evaluation is done on ShapeNetPart.}
    \label{tab:text_encoder_ablation}
\end{table}

\begin{table}[t]
    \centering
    \begin{tabular}{lcc}
        \toprule
        Freezing strategy & mIoU & cIoU \\
        \midrule
        Freeze last block (ours) & \textbf{56.90} & \textbf{53.10} \\
        Freeze last two blocks & 55.70 & 50.93 \\
        Freeze last three blocks & 55.24 & 49.62 \\
        Full encoder frozen & 53.95 & 50.10 \\
        Full fine-tuning & 49.40 & 48.75 \\
        \bottomrule
    \end{tabular}
    \caption{\textbf{Ablation on the freezing strategy during Stage 2.}
    We freeze most of the encoder to preserve Stage 1 visual knowledge and
    avoid destructive interference with the text-alignment objective. Best
    results are obtained by unfreezing only the projection head and the final
    transformer block. Evaluation is done on ShapeNetPart.}
    \label{tab:freezing_ablation}
\end{table}

\paragraph{Ablation on the freezing strategy.}
Stage 1 provides high-quality geometric and semantic priors, and Stage 2 must
align them to text without overriding these learned representations.  
\cref{tab:freezing_ablation} shows that fully fine-tuning the encoder harms
performance, indicating that the text objective alone is not sufficient to
preserve Stage 1 knowledge. Conversely, freezing the entire encoder limits the
ability to adapt to the language space.  
The best trade-off is obtained by freezing all but the last transformer block
and the projection head, which allows for gentle adaptation while preserving
most Stage 1 features. This strategy is used in PatchAlign3D.

\begin{table}[t]
    \centering
    \begin{tabular}{lccc}
        \toprule
        Setting $(k \times G)$ & mIoU & Boundary mIoU & Speed (s) \\
        \midrule
        $64 \times 64$   & 48.1 & 26.03 & 0.40 \\
        $32 \times 128$  & 50.5 & 26.46 & 0.40 \\
        $16 \times 128$  & 49.5 & 28.53 & 0.39 \\
        $16 \times 256$  & \textbf{50.7} & \textbf{29.65} & 0.40 \\
        $8 \times 256$   & 43.7 & 24.89 & 0.39 \\
        $4 \times 512$   & 46.7 & 26.58 & 0.36 \\
        $2 \times 1024$  & 42.0 & 24.88 & 0.36 \\
        \midrule
        Find3D~\cite{ma2025find} & 23.3 & 22.90 & 0.40 \\
        \bottomrule
    \end{tabular}
    \caption{\textbf{Ablation on patch granularity during Stage 2.}
    We vary the patch size $k$ and the number of patches $G$ while keeping the
    overall input size fixed. Intermediate patch sizes provide the best
    trade-off between semantic context and boundary precision, while very small
    patches approach a point-wise regime and lead to a clear drop in
    performance. Evaluation is done on ShapeNetPart.}
    \label{tab:patch_granularity_ablation}
\end{table}

\paragraph{Ablation on patch granularity.}
We further evaluate the impact of the patch partitioning used in Stage 2 in
\cref{tab:patch_granularity_ablation}. PatchAlign3D performs best with
intermediate patch sizes, with the configuration $(k,G)=(16,256)$ achieving the
highest mIoU and boundary mIoU. This confirms that aggregating points into
local patches is important for robust semantic alignment. In contrast, pushing
the model toward a nearly point-wise regime by reducing the patch size to
$k=2$ significantly degrades performance. This supports our hypothesis that
patch-level context helps absorb the noise and ambiguity of pseudo part
annotations while preserving sufficiently precise localization. We also observe
that inference speed remains nearly constant across configurations, indicating
that the gains are not due to increased computational cost. Overall, these
results validate the central design choice of learning language-aligned local
representations at the patch level rather than directly at the point level.

\paragraph{Boundary analysis.}
Although PatchAlign3D predicts labels at the patch level, the model is trained
with fractional multi-label supervision: when a patch overlaps multiple parts,
its target distribution reflects the proportion of points assigned to each
part. This makes boundary patches explicitly informative rather than ambiguous
training failures. To quantify the quality of predictions near part transitions,
we report the boundary mIoU in \cref{tab:patch_granularity_ablation}, computed
on points whose local neighborhoods (k=16) contain mixed semantic labels. The results show a controlled trade-off between spatial granularity and semantic context
rather than a catastrophic collapse near boundaries. In particular, the best
configuration also achieves the highest boundary mIoU, indicating that the
patch representation preserves fine-grained shape details while remaining
robust to noisy supervision.

\begin{table}[t]
    \centering
    \begin{tabular}{lcc}
        \toprule
        Prompt type & mIoU & cIoU \\
        \midrule
        Part-only (base) & 53.1 & 56.9 \\
        Part + category & \textbf{54.1} & \textbf{58.3} \\
        Hard synonyms & 46.2 & 50.9 \\
        \bottomrule
    \end{tabular}
    \caption{\textbf{Prompt sensitivity at inference time.}
    We evaluate the robustness of PatchAlign3D to different textual query
    formulations while keeping the trained model fixed. Adding category context
    slightly improves performance, while even difficult synonyms remain
    competitive. Evaluation is done on ShapeNetPart.}
    \label{tab:prompt_robustness_ablation}
\end{table}

\paragraph{Prompt sensitivity and robustness.}
\cref{tab:prompt_robustness_ablation} evaluates the sensitivity of
PatchAlign3D to the textual formulation used at inference time. The model is
trained once using simple generic part names and then queried with different
prompt variants. We observe that adding category context leads to a small but consistent improvement.
More challenging synonyms lead to lower performance, but remain reasonably
competitive, indicating that the learned patch-level features are not overly
sensitive to a single fixed wording. These results support our claim that
PatchAlign3D can generalize beyond the exact labels seen during training while
still benefiting from informative prompts.

\begin{table}[t]
    \centering
    \begin{tabular}{lc}
        \toprule
        Negative sampling strategy & mIoU \\
        \midrule
        Within-shape negatives only (ours) & \textbf{56.9} \\
        + Cross-shape negatives & 45.3 \\
        \bottomrule
    \end{tabular}
    \caption{\textbf{Effect of cross-shape negatives in Stage 2.}
    Adding negatives from other shapes in the batch significantly degrades
    performance, suggesting that many of these pairs correspond to false
    negatives under incomplete or noisy part annotations. Evaluation is done on ShapeNetPart.}
    \label{tab:cross_shape_negatives}
\end{table}

\paragraph{Ablation on negative sampling.}
In Stage 2, PatchAlign3D uses negatives defined only within each shape.
\cref{tab:cross_shape_negatives} evaluates a hybrid variant that additionally
treats labels from other shapes in the batch as negatives. This modification
substantially degrades performance. We attribute this to the noisy and partial
nature of the pseudo part annotations: semantically identical or closely
related parts across different shapes may be incorrectly treated as negatives,
thereby weakening the alignment objective. These results justify our use of
within-shape negatives only, which provides a more reliable supervision signal
for open-world part understanding.

\section{Additional Qualitative Comparisons}

\begin{figure}[t]
    \centering
    \includegraphics[width=\columnwidth]{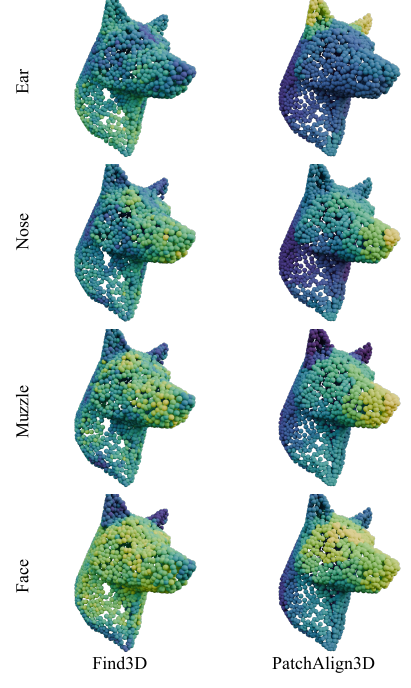}
    \caption{\textbf{Text-to-feature similarity visualization.}
    We compare PatchAlign3D to Find3D by visualizing similarities between a
    textual query (e.g., \emph{``ear''}, \emph{``nose''}) and the dense features
    on a validation point cloud. Yellow indicates higher similarity.
    PatchAlign3D produces sharper and more localized responses, while Find3D
    often shows diffuse signals with weaker semantic localization.}
    \label{fig:qualitative_text}
\end{figure}

\paragraph{Text-to-feature similarity.}
Figure~\ref{fig:qualitative_text} highlights the difference between our
patch-level alignment and Find3D's~\cite{ma2025find} point-wise contrastive learning. Find3D's
responses are often noisy and lack spatial precision, whereas PatchAlign3D
produces clean, well-localized activations that align closely with the queried
part. Related queries (e.g., \emph{``nose''} vs. \emph{``muzzle''}) activate
similar regions, demonstrating semantic consistency in the learned feature
space.

\begin{figure}[t]
    \centering
    \includegraphics[width=\columnwidth]{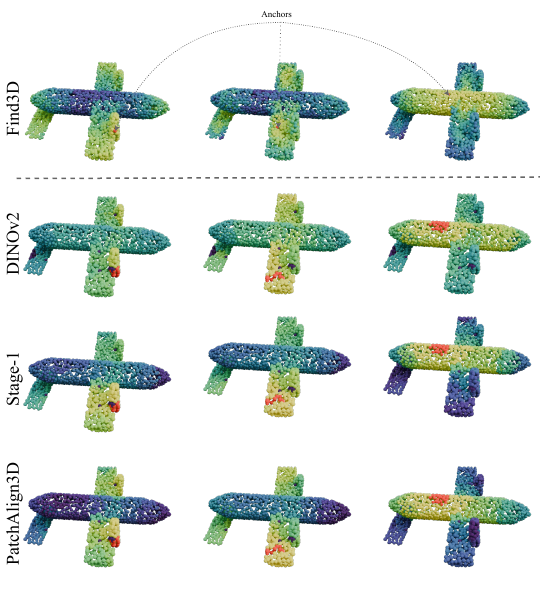}
    \caption{\textbf{Anchor-based feature similarity.}
    For a selected anchor point or patch on a shape (e.g., wing, body, motor),
    we visualize the similarity of all other points/patches to the anchor.
    PatchAlign3D shows stronger geometric coherence than DINOv2, Stage 1, and
    Find3D.}
    \label{fig:qualitative_anchor}
\end{figure}

\paragraph{Anchor-based feature similarity.}
Figure~\ref{fig:qualitative_anchor} provides deeper insight into the structure
of the learned representations. DINOv2~\cite{oquab2023dinov2} features exhibit limited contrast and
lack clear part boundaries. Stage 1 improves geometric coherence but may
preserve symmetries or artifacts from multi-view lifting. Find3D~\cite{ma2025find} captures part
structure but with weaker separation between fine-grained regions.
PatchAlign3D produces the cleanest and most discriminative part clusters,
demonstrating that Stage 2 refines Stage 1 features while preserving geometric
priors.

\section{Zero-shot and Few-shot Keypoint Detection}

In this section, we introduce a zero-shot approach for keypoint detection on 3D shapes. Compared to semantic segmentation, reasoning at the point level on visual data is challenging because it requires precise localization capabilities, which can be problematic even for advanced models like GPT-4o~\cite{achiam2023gpt} and PaliGemma 2~\cite{steiner2024paligemma}.

Traditionally, 3D keypoint detection relies heavily on annotated 3D datasets and extensive supervised training, which limits its scalability and its applicability to new categories or domains. In contrast, the zero-shot method takes advantage of the rich knowledge embedded within language models. Specifically, we show that part-level annotations used to train 3D encoders can be employed to detect salient keypoints on 3D models without requiring any ground truth labels or supervision.

We evaluate our method using the KeypointNet dataset, which provides dense annotations and text prompts for keypoints. Our evaluation strategy and baselines are based on the method described by 
\citet{gong2025zerokey}. This evaluation computes the Intersection over Union (IoU) between predicted keypoints and ground-truth keypoints from the KeypointNet~\cite{you2020keypointnet} dataset, using different distance thresholds. A match is counted when the geodesic distance between a ground-truth keypoint and a predicted keypoint is less than the specified threshold.

\begin{table*}[t]
\centering
\begin{tabular}{lccccccccccc}
\toprule
\textbf{Method} & \textbf{0.001} & \textbf{0.01} & \textbf{0.02} & \textbf{0.03} & \textbf{0.04} & \textbf{0.05} & \textbf{0.06} & \textbf{0.07} & \textbf{0.08} & \textbf{0.09} & \textbf{0.10} \\
\midrule
HARRIS-3D~\cite{sipiran2011harris} & 0.15 & 0.76 & 2.19 & 3.96 & 6.16 & 8.88 & 11.91 & 15.13 & 18.63 & 22.10 & 25.69 \\
SIFT-3D~\cite{rister2017volumetric} & 0.29 & 1.05 & 2.62 & 4.83 & 6.95 & 9.42 & 12.38 & 15.65 & 19.39 & 22.71 & 26.15 \\
ISS~\cite{zhong2009intrinsic} & 0.32 & 1.19 & 2.79 & 4.76 & 6.93 & 9.40 & 12.04 & 15.10 & 18.32 & 22.08 & 25.87 \\
USIP~\cite{li2019usip} & 0.83 & 1.70 & 3.25 & 5.24 & 8.07 & 11.15 & 15.98 & 20.56 & 25.36 & 30.16 & 34.77 \\
D3FEAT~\cite{bai2020d3feat} & 2.36 & 3.86 & 7.82 & 12.77 & 18.53 & 25.02 & 31.14 & 36.65 & 41.74 & 46.33 & 50.52 \\
UKPGAN~\cite{you2022ukpgan} & 3.95 & 6.54 & 12.77 & 18.22 & 26.45 & 35.32 & 40.28 & 34.42 & 42.65 & 46.05 & 46.49 \\
FSKD~\cite{attaiki2022ncp} & 7.00 & 7.94 & 11.17 & 17.67 & 23.99 & 31.14 & 38.14 & 43.97 & 49.32 & 53.87 & 57.05 \\
B2-3D~\cite{wimmer2024back} & 6.20 & 11.87 & 19.63 & 27.65 & 31.14 & 34.64 & 38.86 & 41.95 & 44.77 & 46.69 & 49.25 \\
ULIP-2~\cite{xue2023ulip} & 2.00 & 3.85 & 7.09 & 9.31 & 11.22 & 13.11 & 15.23 & 17.57 & 19.95 & 22.34 & 24.88 \\
\specialrule{0.5pt}{1pt}{1pt} 
\rowcolor{blue!10}
\begin{tabular}{@{}l@{}} PatchAlign3D \\ - Few Shot (Ours) \end{tabular}
& \textbf{7.80} & \textbf{12.16} & \textbf{21.63} & \textbf{30.54} & \textbf{37.48} & \textbf{43.61} & \textbf{48.96} & \textbf{53.51} & \textbf{57.46} & \textbf{60.97} & \textbf{64.07} \\
\midrule
PaliGemma 2~\cite{steiner2024paligemma} & 0.00 & 0.34 & 1.03 & 2.98 & 4.93 & 7.00 & 6.62 & 8.17 & 8.92 & 11.49 & 11.56 \\
RedCircle~\cite{shtedritski2023does} & 0.21 & 0.34 & 0.64 & 1.16 & 1.90 & 3.04 & 4.81 & 7.55 & 11.06 & 14.92 & 18.50 \\
GPT-4o~\cite{achiam2023gpt} & 0.28 & 0.38 & 1.04 & 2.11 & 4.79 & 6.58 & 8.48 & 10.09 & 14.03 & 17.03 & 17.85 \\
CLIP-DINoiser~\cite{wysoczanska2024clip} & 0.73 & 1.41 & 3.00 & 4.94 & 7.31 & 9.81 & 12.66 & 15.52 & 18.52 & 21.76 & 25.56 \\
\specialrule{0.5pt}{1pt}{1pt} 
\rowcolor{green!12}
\begin{tabular}{@{}l@{}} PatchAlign3D \\ - Zero Shot (Ours) \end{tabular}
& \textbf{2.21} & \textbf{3.94} & \textbf{9.48} & \textbf{16.13} & \textbf{19.88} & \textbf{22.04} & \textbf{23.32} & \textbf{25.23} & \textbf{27.85} & \textbf{31.44} & \textbf{32.88} \\
\bottomrule
\end{tabular}
\caption{Comparison of IoU between the predicted and ground-truth keypoints from KeypointNet using different methods across various geodesic distance thresholds. The \textbf{\colorbox{blue!10}{blue}} text indicates the best few-shot methods, while the \textbf{\colorbox{green!10}{green}} text highlights the best zero-shot methods.}
\label{tab:comparison_suppl}
\end{table*}

\begin{figure}[t]
    \vspace{-2em}
    \centering
    \includegraphics[width=\columnwidth]{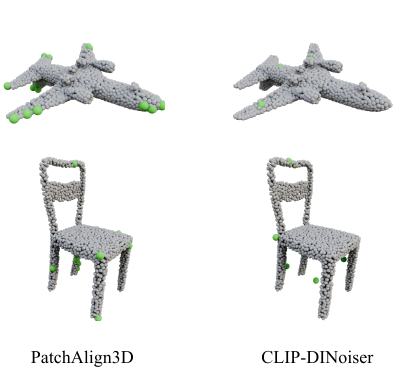}
    \caption{\textbf{Visualization of keypoints detected in zero-shot keypoint detection.} In these experiments, the input to our method is a point cloud containing 2048 points. The detected keypoints given a text prompt are highlighted as larger green dots.}
    \label{fig:qualitative_kp}
\end{figure}

Among all the baselines, both RedCircle~\cite{shtedritski2023does} and CLIP-DINOiser~\cite{wysoczanska2024clip} utilize text alignment from CLIP, allowing for the querying of keypoints using text in a zero-shot setting. However, both methods are multiview-based and do not incorporate explicit 3D modeling. Our approach also leverages CLIP for feature alignment with text, but it benefits from a 3D point encoder. This makes RedCircle and CLIP-DINOiser ideal baselines for comparison with our method.

Our evaluation of KeypointNet~\cite{you2020keypointnet} demonstrates (refer to Fig. \ref{fig:qualitative_kp} and Tab. \ref{tab:comparison_suppl}) that our text-aligned local feature significantly outperforms other baselines, including RedCircle, CLIP-DINOiser, and even GPT-4o, across all distance thresholds. Additionally, in few-shot settings where text alignment is not required, our patch-adopted feature significantly surpasses the globally adopted ULIP-2~\cite{xue2023ulip} features and the multi-view aggregated features from B2-3D.

This provides strong evidence that our language-aligned local features are not only more semantically meaningful but also serve as better geometry descriptors compared to globally supervised features. Furthermore, our method achieves IoU levels comparable to those of supervised methods specifically designed for this dataset, such as B2-3D~\cite{wimmer2024back} and FSKD~\cite{attaiki2022ncp}. These results emphasize that our feature improved point-level understanding of both text semantics and geometry through fine-grained patch alignment.

\begin{table}[t]
    \centering
    \begin{tabular}{lccc}
        \toprule
        Method & IoU@0.01 & IoU@0.05 & IoU@0.10 \\
        \midrule
        B2-3D~\cite{wimmer2024back} & 6.20 & 31.14 & 46.69 \\
        ULIP-2~\cite{xue2023ulip} & 2.00 & 11.22 & 22.34 \\
        PatchAlign3D (ours) & 7.80 & 37.48 & 60.97 \\
        StablePoints & 3.66 & 16.58 & 34.89 \\
        ZeroKey~\cite{gong2025zerokey} & \textbf{13.16} & 56.60 & 79.43 \\
        Ours + ZeroKey & 10.42 & \textbf{58.87} & \textbf{81.27} \\
        \bottomrule
    \end{tabular}
    \caption{\textbf{Integration of PatchAlign3D into ZeroKey.}
    We incorporate our patch-level features into the ZeroKey pipeline by
    modulating its soft-voting weights with patch--text similarities and by
    injecting local feature cues into the clustering stage. The combination
    improves performance at medium and large geodesic thresholds, indicating
    that PatchAlign3D provides complementary fine-grained geometric cues for
    zero-shot keypoint reasoning.}
    \label{tab:zerokey_integration}
\end{table}

\paragraph{Integration with ZeroKey.}
We further evaluate whether PatchAlign3D can benefit recent zero-shot
keypoint detection methods. To this end, we incorporate our patch-level
features into the ZeroKey pipeline by modulating its soft-voting scores with
patch-to-text similarities and by injecting local feature cues into the
clustering stage. As shown in \cref{tab:zerokey_integration}, this hybrid
variant improves over ZeroKey at medium and large geodesic thresholds,
suggesting that PatchAlign3D provides complementary local geometric
information. These results indicate that our features are not only useful for
part segmentation, but can also strengthen other fine-grained open-world 3D
reasoning tasks. \fi

\end{document}